%% file: hvrp_main.tex
\documentclass[a4paper, 11pt]{article}

\usepackage[margin=1in]{geometry}
\usepackage{setspace}
\usepackage{microtype}

\usepackage{amsmath} \allowdisplaybreaks
\usepackage{amssymb}
\usepackage{amsthm}
\usepackage{mathtools}

\usepackage{natbib}
\usepackage{bibentry}

\usepackage{booktabs} %
\usepackage{appendix}
\usepackage[normalem]{ulem} %
\usepackage{mathtools}

\usepackage[hyphens]{url}
\usepackage[bookmarks=false,hidelinks]{hyperref}

\usepackage{tikz}
\usetikzlibrary{automata,calc,trees,positioning,arrows,chains,shapes.geometric,%
decorations.pathreplacing,decorations.pathmorphing,shapes,%
matrix,shapes.symbols,plotmarks,decorations.markings,shadows}

\usepackage{pgf}
\pgfdeclarelayer{background}
\pgfdeclarelayer{foreground}
\pgfsetlayers{background,main,foreground}

\usepackage{authblk}

\usepackage{graphics,graphicx,subfig,float}
\usepackage{multirow}
\usepackage{amsfonts, bbm}
\usepackage{algorithm}
\usepackage{algpseudocode}
 \usepackage{pdflscape}
\usepackage{longtable}

\newcommand{\email}[1]{{\href{mailto:#1}{\nolinkurl{#1}}}}

\usepackage{bm}

\DeclareMathSymbol{\Gamma}{\mathord}{operators}{"00}
\DeclareMathSymbol{\Delta}{\mathord}{operators}{"01}
\DeclareMathSymbol{\Theta}{\mathord}{operators}{"02}
\DeclareMathSymbol{\Lambda}{\mathord}{operators}{"03}
\DeclareMathSymbol{\Xi}{\mathord}{operators}{"04}
\DeclareMathSymbol{\Pi}{\mathord}{operators}{"05}
\DeclareMathSymbol{\Sigma}{\mathord}{operators}{"06}
\DeclareMathSymbol{\Upsilon}{\mathord}{operators}{"07}
\DeclareMathSymbol{\Phi}{\mathord}{operators}{"08}
\DeclareMathSymbol{\Psi}{\mathord}{operators}{"09}
\DeclareMathSymbol{\Omega}{\mathord}{operators}{"0A}

\usepackage{xcolor}

\usepackage{etoolbox}
\makeatletter
\def\do#1{\@namedef{#1c}{\ensuremath{\mathcal{#1}}}}
\docsvlist{A,B,C,D,E,F,G,H,I,J,K,L,M,N,O,P,Q,R,S,T,U,V,W,X,Y,Z}
\makeatother

\renewcommand{\bar}[1]{\mkern 1.5mu\overline{\mkern-1.5mu#1\mkern-1.5mu}\mkern 1.5mu}

\newcommand{\RNum}[1]{\uppercase\expandafter{\romannumeral #1\relax}}

\title{Genetic Algorithms with Neural Cost Predictor for Solving Hierarchical Vehicle Routing Problems}

\author[1]{Abhay Sobhanan}
\author[2,3]{Junyoung Park}
\author[2,3]{Jinkyoo Park}
\author[2,3]{Changhyun Kwon}
\affil[1]{University of South Florida, Tampa, FL 33620, U.S.A.}
\affil[2]{KAIST, Daejeon, 34141, Republic of Korea}
\affil[3]{OMELET, Daejeon, 34051, Republic of Korea}

\date{}

\begin{document}

\maketitle

\begin{abstract}
When vehicle routing decisions are intertwined with higher-level decisions, the resulting optimization problems pose significant challenges for computation. Examples are the multi-depot vehicle routing problem (MDVRP), where customers are assigned to depots before delivery, and the capacitated location routing problem (CLRP), where the locations of depots should be determined first. A simple and straightforward approach for such hierarchical problems would be to separate the higher-level decisions from the complicated vehicle routing decisions. For each higher-level decision candidate, we may evaluate the underlying vehicle routing problems to assess the candidate. As this approach requires solving vehicle routing problems multiple times, it has been regarded as impractical in most cases.  We propose a novel deep-learning-based approach called Genetic Algorithm with Neural Cost Predictor (GANCP) to tackle the challenge and simplify algorithm developments. For each higher-level decision candidate, we predict the objective function values of the underlying vehicle routing problems using a pre-trained graph neural network without actually solving the routing problems. In particular, our proposed neural network learns the objective values of the HGS-CVRP open-source package that solves capacitated vehicle routing problems. Our numerical experiments show that this simplified approach is effective and efficient in generating high-quality solutions for both MDVRP and CLRP and has the potential to expedite algorithm developments for complicated hierarchical problems. We provide computational results evaluated in the standard benchmark instances used in the literature.
\end{abstract}

\textbf{Funding:} This research was funded by the Korean government (MSIT) through the National Research Foundation of Korea (NRF) grant RS-2023-00259550.

\textbf{Key words:} multi-depot vehicle routing problem, genetic algorithm, deep learning, cost prediction

\input{hvrp_contents.tex}

\section*{Acknowledgement} 
This research was supported by the Korean government (MSIT) through the National Research Foundation of Korea (NRF) grant RS-2023-00259550.
Disclaimer: Any opinions, findings, conclusions, or recommendations expressed in this material are those of the authors.

\bibliographystyle{ormsv080-ck}
\bibliography{hvrp_ref}

\newpage
\begin{appendices}

\section{Capacitated Location Routing Problem}

\input{hvrp_appendix.tex}

\begin{landscape}
\section{Detailed Results of Computational Experiments} \label{sec:detailed_results}

\fontsize{5.8}{8}\selectfont
\input{tables/hvrp_table_T.tex}

\newpage
\footnotesize
\input{tables/hvrp_table_O.tex}

\newpage
\input{tables/hvrp_table_D.tex}

\newpage
\input{tables/hvrp_table_D_tuned.tex}

\newpage
\scriptsize
\input{tables/hvrp_table_CLRP_detailed.tex}

\end{landscape}

\normalsize
\end{appendices}

\end{document}

%% file: hvrp_contents.tex
\section{Introduction}

Vehicle Routing Problems (VRPs) have garnered significant attention over the past few decades due to their crucial role in last-mile delivery logistics. 
\citet{capgemini} reports that last-mile delivery accounts for 41\% of the supply chain expenditures.
As a direct extension of the Traveling Salesman Problem (TSP), the Capacitated Vehicle Routing Problem (CVRP) forms a fundamental class of VRPs and serves as the basis for addressing many last-mile delivery challenges. 
In many real-life routing scenarios, the VRP under consideration often includes additional constraints, parameters, or assumptions. 
Such problems are referred to as VRP variants or rich VRPs. 

Among the many VRP variants, we focus on hierarchical VRPs (HVRPs), where the higher-level decisions impact the lower-level decisions. 
Examples include two widely used VRP variants called the Multi-Depot VRP (MDVRP) and the Capacitated Location Routing Problem (CLRP). 
In essence, the MDVRP involves the higher-level task of clustering customers with order-fulfillment centers, from which the respective demands are fulfilled. 
Clustering decomposes the original problem into multiple CVRPs.
Evidently, the MDVRP is at least as difficult as the CVRP, and thus an $\mathcal{NP}$-Hard problem. 
In MDVRPs, the higher-level decisions involve assigning each customer to a depot, whereas the lower-level decisions focus on routing vehicles from each depot to its assigned customers.
As a generalization of the MDVRP, the CLRP requires additional higher-level decisions that determine which depots should be opened from the available candidate locations. 

The main challenge in obtaining real-time solutions to combinatorial optimization (CO) problems lies in their computational complexity. 
Although heuristics tackle this issue for relatively small and medium-sized problems, the search space for good solutions grows exponentially with the problem size. 
Many logistics problems in the practical settings encompass several hundred or even thousands of customers \citep{qi2012spatiotemporal}.
The MDVRP is a fundamental optimization problem in many real-world transportation problems, including parcel delivery, waste collection, mobile healthcare routing, and field service management. 
Some of these routing scenarios involve solving large networks comprising thousands of locations and need to be tackled on a daily basis, such as waste collection and e-commerce order fulfillment \citep{arnold2019efficiently}.
In waste collection, numerous trucks dispatched from various waste disposal plants (depots) are tasked with collecting waste from tens of thousands of customers.
In 2022, a prominent e-commerce firm \textit{Amazon Logistics} processed 4.79 billion U.S. delivery orders \citep{capitalone}, equating to approximately 9,116 orders per minute. 
These examples underscore the magnitude and complexity of large-scale vehicle routing faced by real-world businesses and the need for fast operational decisions.

As another example of an HVRP, consider grocery deliveries by an organization that operates a chain of grocery stores or hypermarkets across an urban area. 
If a customer's delivery demands can be satisfied from multiple stores, the company needs to decide which store fulfills the customer demand more profitably. 
With several incoming orders from different geographical locations and a limited number of delivery agents available at each store, the organization needs to determine the optimal assignment of each customer to a store so that the overall transportation cost is minimized. 
It is evident that allocating one or more customers to their nearest store, disregarding other factors such as the geographical distribution of demands and the delivery capacities, may lead to suboptimal or even infeasible solutions.
Presently, a prevalent method to address such large-scale HVRPs involves a cluster first and route second approach. However, employing a fixed strategy for higher-level decisions may result in increased operational costs.

The objective function value of an MDVRP with customer-depot assignments is calculated by adding the optimal costs of all decomposed CVRPs. 
For problems that can be decomposed into “simpler” subproblems, a convenient solution method is to find the optimal decomposition. 
However, recall that an MDVRP decomposition leads to multiple CVRPs that are still $\mathcal{NP}$-Hard. 
To evaluate a decomposition, the only requirement is the optimal routing cost of each subproblem without the need for routing decisions. 
In this research, we train a deep neural network (NN) model to predict the optimal cost of each CVRP instance arising from the decomposition of an HVRP, without actually solving the CVRPs. 
The predicted cost can be leveraged as an approximate measure of the decomposition strategy's effectiveness.
With the help of a Genetic Algorithm (GA), we improve the clustering of customers with depots. 
The fitness cost of an assignment is calculated from the sum of subproblems' cost predictions and the distinctiveness of the decomposition compared to other assignments. 
We refer to this approach as the Genetic Algorithm with Neural Cost Predictor ($\text{GANCP}$).

Our research focuses on the MDVRP as a representative case of the broader HVRP category, with additional experiments on the CLRP detailed in the Appendix.
The numerical experiments demonstrate the effectiveness of $\text{GANCP}$ in obtaining high-quality solutions while offering significant computational advantages.
At the final stage of the proposed method, we use the HGS-CVRP solver \citep{vidal2022hybrid} to obtain an actual routing solution. 
Therefore, we refer to our complete end-to-end solution framework with the use of the subproblem solver as $\text{GANCP}^+$. 
We compare the test results with the performance of an open-source generic VRP solver, the Vehicle Routing Open-source Optimization Machine (VROOM) \citep{vroom}. 
In many VRP variants, VROOM is known to produce high-quality solutions that are close to the best-known solutions reported in the literature.
We analyze the performance of the cost prediction model using three different train data generation procedures. 
We also showcase the transferability of our heuristic to instances that are different from the trained distribution and size.

We believe that the proposed $\text{GANCP}^+$ has the potential to tackle many challenging hierarchical VRPs that arise in the real world for the following reasons.
First, the trained NN can predict the cost of each candidate quickly. 
The biggest challenge in the decomposition approach for hierarchical VRPs is the time to solve subproblems.
In our approach, we use the neural network to quickly generate a cost prediction without solving the problem.
Second, our proposed NN can learn the cost values from standard VRP solvers for a wide range of problem sizes. 
Fundamental building blocks in hierarchical VRPs are well-studied problems such as CVRP, for which several outstanding VRP solvers exist. 
Therefore, we can train the NN for each standard VRP class and use it for various HVRPs that use the same VRP class of problems as the subproblems. 
This can significantly reduce algorithm development time, as we can focus on handling non-standard modeling components within the GA framework without considering the routing solutions.
Third, our proposed NN model can provide a robust prediction capability for the same VRP class within different problem contexts. 
We demonstrate the transferability of our trained NN model from the MDVRP to the CLRP by experimenting with multiple benchmark datasets.

Our \emph{goal} in this paper is, therefore, to show that the proposed approach can \emph{simplify} algorithm development processes for complicated hierarchical vehicle routing problems and produce quality solutions in a short amount of time rather than to develop an algorithm that can produce the best solutions and outperform all existing algorithms. 
We will focus on illustrating this main point throughout the paper and will use MDVRP and CLRP as examples. 

The remainder of this paper is organized as follows.
In Section \ref{sec:lit_review}, we discuss the relevant literature on the CVRP methodologies involving reinforcement learning (RL), machine learning (ML), heuristics, and cost estimation techniques, followed by a brief review of related literature on MDVRP and CLRP.
Section \ref{sec:hvrp} describes the HVRPs in general and focuses on the decomposition of the MDVRP to VRPs based on the mathematical formulation.
The neural network prediction model and training details are discussed in Section \ref{sec:nn}.
Section \ref{sec:mdvrp} explains our MDVRP solution methodology that incorporates the NN model architecture and the heuristics in detail. 
We discuss and evaluate the performance of the NN model using three different training procedures in Section \ref{sec:comp_exp1}.
We further demonstrate the effectiveness of our proposed heuristic on multiple MDVRP instances, including out-of-distribution and out-of-range instances.
The paper finally concludes with a discussion of relevant and related research directions in optimization using supervised learning in Section \ref{sec:conclusion}.
In addition, we demonstrate the transferability of our approach with a simple extension of the MDVRP methodology for the CLRP in Section \ref{sec:clrp} and examine the results of the CLRP experiments in Section \ref{sec:comp_exp2}.

\section{Literature Review} \label{sec:lit_review}

First, we review the relevant literature on VRPs that use deep RL as a solution methodology. 
Subsequently, we explore the supervised learning approaches applied to VRPs. 
We further discuss pertinent approximation techniques utilized in the vehicle routing literature for optimal cost estimations.
This is followed by a review of relevant research on solution methods for the MDVRP and the CLRP.
Finally, we highlight our contributions to the existing research gap in this domain.

\subsection{Deep Reinforcement Learning Approaches for VRPs}

Over the past few years, deep reinforcement learning has been demonstrating notable success in solving combinatorial optimization problems, especially the TSP and the CVRP.
\citet{bello2016neural} introduces an effective deep RL framework to train and solve the TSP using pointer networks \citep{vinyals2015pointer}. 
The problem is modeled as a Markov Decision Process and uses the pointer network for encoding the inputs, which are then learned using policy gradients. 
\citet{nazari2018reinforcement} extends this work for VRPs without a Recurrent Neural Network for the input encoding. 
Unlike Natural Language Processing, the order of the input sequence---the customer locations---can be treated independently in the VRP.
While the inclusion of a pointer encoder network does not negatively impact the final accuracy, it delays learning due to added model complexity. 
\citet{kool2018attention} yields better performance for CVRPs using an architecture for solution construction with multiple multi-head attention layers, proposed in \citet{vaswani2017attention}, to capture the node dependencies, and using the REINFORCE algorithm to train the model. 
With consideration of symmetries in the solutions of a CO problem, \citet{kwon2020pomo} designs the Policy Optimization with Multiple Optima (POMO) network with a modified REINFORCE algorithm to explore the solution space more effectively. 
This improves the results in terms of both performance gap and inference time compared to the existing architectures.
Recent studies have also addressed different VRP variants using deep RL frameworks \citep{bogyrbayeva2023deep, park2023learn}. 
All of these approaches are end-to-end solution approaches that rely solely on neural networks to solve the problem. 

Traditional end-to-end RL frameworks for CO solution construction have limitations concerning the scope and size of the problems they can effectively address. 
Another active research focus is on the use of RL as an enhancement heuristic for problem partitions or when an initial solution is known. 
\cite{Lu2020A} utilize an RL agent as a heuristic controller to apply intra-route and inter-route improvement operators to an initial randomly generated solution and obtain high-quality CVRP solutions.
In a similar study, \cite{wu2021learning} proposes an RL model that employs a single improvement operator instead of multiple operations in combination. 
The authors compare operators such as 2-opt, node swap, and relocation and illustrate that 2-opt yields better results.
\citet{zong2022rbg} proposes a Rewriting-by-Generating framework for solving large-scale VRPs hierarchically.
Here, the Rewriter agent partitions the customers into independent regions, and the Generator agent solves the routing problem in each region.
This method outperforms the HGS-CVRP solver by \citet{vidal2022hybrid} for large-scale problems, where the hyperparameters are fine-tuned for small to medium-size CVRP problems. 
However, RL requires higher computational time for training and inference compared to supervised machine learning and faces challenges in exploration due to the iterative nature of reinforcement learning with the environment.

\subsection{Supervised Learning Approaches for VRPs}

Multiple supervised learning frameworks exist for solving the VRP through the integration of exact methods, heuristics, or predictions of subproblem costs. 
The integration of ML with exact optimization techniques proves considerable promise and is currently a subject of active research.
For example, \cite{furian2021machine} solves VRPs using a branch-and-price method where ML models are used to predict the decision variables and branching scores. 
Another interesting study is \cite{morabit2021machine}, where a Graph Neural Network (GNN) acts as a column selector to accelerate column generation for the VRP with time windows. 
\citet{kim2024neural} further demonstrates that GNN can be leveraged to solve the separation problems in cutting plane methods, particularly for large-scale instances.

A few recent studies utilize supervised machine learning to learn potential cost improvements for a subproblem and iteratively improve the solution. 
\citet{li2021learning} proposes a method that iteratively improves the solution to large-scale VRPs by learning to select subproblems and estimating the corresponding subsolution costs using a transformer architecture. 
Using a combination of the subproblem selector and cost estimation, the problem can be partitioned into smaller problems that are easily solved.
\citet{kim2023learning} presents a neural cross-exchange operator to solve multiple VRP variants based on cost decrement predictions and demonstrates its computational benefits against the traditional operator. 
In a recent study, \cite{varol2024neural} proposed a neural network tour length estimator for a very large-scale Generalized TSP, incorporating instance-specific features to enhance estimation accuracy.
The authors employ a similar approach to ours, integrating the neural network with a genetic algorithm in the solution framework. 
However, their estimator performs well only on specific problem sizes, in contrast to our NN model.
The results of these studies showcase the efficacy of supervised learning in addressing large-scale routing problems. 
This stream of research motivates our proposed methodology.

\subsection{Optimal Cost Estimations of VRPs}  %

Numerous studies aim to accurately estimate optimal routing costs based on the topological distribution of customers, demands, and vehicle capacities. 
These studies typically utilize two methodologies: Continuous Approximation (CA) and Regression. 
CA, as an alternative to discrete models, employs smooth density functions to represent the network and is particularly effective for larger instances. 
CA techniques find application in estimating optimal transportation costs across various VRP variants, including emission minimization VRP \citep{saberi2012continuous}, school bus routing \citep{ellegood2015continuous}, hub location routing problems \citep{ghaffarinasab2018continuous}, dynamic multiple TSP \citep{garn2021balanced}, and same-day delivery \citep{banerjee2022fleet, stroh2022tactical}. 
A detailed review of the CA approaches used in freight distribution is available in \citet{franceschetti2017continuous}.

Among analytical expressions, the Daganzo-approximation \citep{robust1990implementing} is widely recognized for estimating the optimal CVRP distance given by following the formula:
\begin{equation}
    \text{CVRP cost} \approx \Big(0.9+ \frac{kN}{C^2} \Big)\sqrt{AN} \label{eq:dist_daganzo}
\end{equation}
where $A$ represents the bounded area within which $N$ number of customers are distributed, $k$ is an area shape constant, and $C$ is the maximum number of customers a vehicle can serve.
\cite{figliozzi2008planning} improves the accuracy of cost approximation using linear regression to calculate the optimal cost based on a few critical instance features as follows:
\begin{equation}
    \text{CVRP cost} \approx a_1 \frac{N-M}{N} \sqrt{AN} + a_2 \frac{A}{N} + a_3 M \label{eq:dist_approx_2008}
\end{equation}
where $M$ is the number of vehicles, and the parameters $a_1, a_2$ and $a_3$ are determined through regression. 
\citet{nicola2019total} and \citet{akkerman2022distance} introduce linear distance approximation models and discuss the importance of critical feature selection in achieving accurate cost estimation.
Both studies perform experiments that cover cases with different distributions and multiple VRP variants. 
\citet{kou2023improved} employ linear regression to estimate the optimal cost of CVRPs based on the objective values of multiple Clarke-Wright heuristic solutions. 
To train each instance, the authors utilize mean, standard deviation, and the minimum tour length of 1,000 solutions as predictors to fit a regression model. 
However, their test results rely on a dataset where the majority of instances have exactly 100 customers. 

CA and regression models require careful adaptation for each problem type with the recognition of critical instance features. 
Further, they do not generalize well when confronted with varying problem sizes and data distributions.
We employ a graph neural network to capture both the topological and demand features of the customers. 
This approach eliminates the need for focused feature engineering across different data types, enabling highly accurate predictions of optimal costs across varying problem sizes and data distributions.

\subsection{Optimization Approaches for MDVRP and CLRP}

Multiple studies use exact methods to solve the MDVRP, such as \cite{laporte1988solving}, \cite{contardo2014new}, and \cite{baldacci2009unified}. 
Similarly, for the study of the CLRP using exact methods, one may refer to \cite{baldacci2011exact}, \cite{belenguer2011branch} and \cite{contardo2014exact}.
As a consequence of the  $\mathcal{NP}$-hardness of these problems, efficient heuristics are critical to the development of industrial routing solutions.

\citet{cordeau1997tabu} proposes a notable Tabu Search (TS) heuristic to solve the MDVRP, the Periodic VRP, and the Periodic TSP. 
A few years later, genetic algorithms emerged as one of the most effective heuristics for vehicle routing. 
\citet{vidal2012hybrid} proposes a fast and efficient hybrid genetic algorithm that solves the MDVRP and the Periodic VRP. 
The algorithm differentiates neighboring solutions using a diversity factor in the fitness cost.
In addition, a pool of infeasible solutions is preserved for diversification and better search space. \citet{vidal2014implicit} further improves the results where the depot, vehicle, and first customer visited on each route are optimally determined using a dynamic programming methodology.
\citet{sadati2021efficient} presents a Variable Neighborhood Search algorithm that solves the MDVRP variants with a tabu-shaking mechanism to improve the search trajectory.
End-to-end RL frameworks for MDVRP are presented in \citet{zou2022improved} based on an improved transformer model and in \citet{zhang2023graph} using Graph Attention Networks. 

One of the most effective existing algorithms for solving the large-scale CLRP is by \citet{schneider2019large}, which explores multiple depot configurations using a Tree-Based Search Algorithm (TBSA). For a given configuration, an MDVRP is solved at the routing phase using a granular tabu search. The paper also discusses different variants of the algorithm with trade-offs in computational time and solution quality. 
More recently, \citet{akpunar2021hybrid} proposed a hybrid adaptive large neighborhood search method that demonstrates improved benchmark performance in a few CLRP benchmark instances. 

Our $\text{GANCP}^+$ heuristic is empowered with a supervised learning model for cost estimation, which requires less training time and is, more importantly, applicable to distinct and large problem sizes, producing near-optimal results. 
To the best of our knowledge, this paper is the first to address solving large-scale hierarchical VRPs using a deep learning architecture to evaluate the subproblem decompositions.

\section{Hierarchical Vehicle Routing Problems} \label{sec:hvrp}

Hierarchical vehicle routing problems (HVRPs) involve decisions made at multiple levels, often executed sequentially, to obtain an optimal routing solution. 
In a large or complex HVRP, decisions made at higher levels can decompose the problem into smaller subproblems. 
A good solution strategy involves identifying effective higher-level decisions that optimize the objective function while considering the original problem constraints.
Higher-level decisions influence subsequent decisions, solution quality, and the optimal objective cost of the original problem. 
Lower-level decisions include detailed vehicle routing and schedule plans to complete deliveries or services designated to each vehicle. 

As the HVRPs require decisions in addition to and including vehicle routing, it is computationally challenging to solve such problems in many real-life scenarios.
This paper examines two prevalent HVRP variants: the MDVRP and the CLRP, with the latter analyzed in the Appendix.
Due to the complexity of the HVRPs, multiple techniques and their combination are often used to solve large instances.
A straightforward approach is to focus on the method to find effective higher-level decisions and solve the corresponding subproblems using an existing efficient algorithm or a proven solver. 
The higher-level decisions decompose the MDVRPs and the CLRPs into subsequent CVRPs, where the total solution cost of the subproblems determines the quality of the higher-level decisions. 
Since the CVRPs are well-researched, we focus on a heuristic method to find the higher-level decisions for the HVRPs. 

In the remainder of this section, we first discuss the arc-based flow formulation of the MDVRP to ensure mathematical clarity. 
Following this, we demonstrate how employing a decomposition approach yields CVRP subproblems that reduce the complexity of the original problem.
Later, we provide a discussion regarding the generalizability of the hierarchical decomposition approach for solving HVRPs.

\subsection{Multi-Depot Vehicle Routing Problem}

Consider a complete and undirected graph $\Gc = (\Vc,\Ec)$, where $\Vc$ represents the set of all nodes and $\Ec$ denotes the set of all edges connecting nodes in $\Vc$.
Let $\Dc \subset \Vc$ be the set of depots, and $\Cc \subset \Vc$ be the set of customers, such that $\Vc = \Dc \cup \Cc$. 
Furthermore, let $\Kc_d$ be the set of vehicles available for order-fulfillment at depot $d \in \Dc$, and $\Kc = \sum_{d \in \Dc} \Kc_d$ is the set of all vehicles.
Each vehicle $k \in \Kc$ has a capacity $Q_k$, and customer $i \in \Cc$ has a positive demand $q_i$ such that $q_i \leq Q_\text{min}$, where $Q_\text{min}$ is the minimum vehicle capacity among the fleet of vehicles. 
Note that a vehicle that starts from a depot must return to the same depot after completing its route.

We examine the mathematical model for MDVRP, originally proposed by \citet{golden1977implementing} and later improved by \citet{ramos2020multi} through the distinct partitioning of vehicles among depots.
Let the binary integer decision variable $x_{ijk}$ equal 1 if vehicle $k$ traverses the edge $(i, j)$ in a feasible solution, and 0 otherwise.
The traversal cost for the edge $(i,j)$ is denoted as $c_{ij}$.
The problem is then formulated as follows:
\begin{align}
    \min\ & \sum_{i \in \Vc} \sum_{j \in \Vc} \sum_{k \in \Kc} c_{ij} x_{ijk} & \label{eq:mdvrp_obj} \\
    \text{s.t.}\ & \sum_{i \in \Vc} \sum_{k \in \Kc} x_{ijk} = 1 & \forall j \in \Cc \label{eq:mdvrp_const1} \\
    & \sum_{i \in \Vc} x_{ihk} - \sum_{j \in \Vc} x_{hjk} = 0 & \forall k \in \Kc; h \in \Vc \label{eq:mdvrp_const3} \\
    & \sum_{i \in \Cc} \sum_{j \in \Vc} q_i x_{ijk} \leq Q_k & \forall k \in \Kc \label{eq:mdvrp_const4} \\
    & \sum_{i \in \Cc} x_{ijk} \leq 1 & \forall j \in \Dc; k \in \Kc_j \label{eq:mdvrp_const6} \\
    & \sum_{i \in \Sc} \sum_{j \in \Sc} x_{ijk} \leq |\Sc|-1 & \forall \Sc \subseteq \Cc; |\Sc| \geq 2; k \in \Kc \label{eq:mdvrp_const7} \\
    & \sum_{j \in \Dc} \sum_{k \in K} x_{ijk} = 0 & \forall i \in \Dc \label{eq:mdvrp_const8} \\ %
    & \sum_{i \in \Cc} x_{ijk} = 0 & \forall j \in \Dc; k \in \Kc \setminus \Kc_j \label{eq:mdvrp_const9} \\
    & x_{ijk} \in \{0, 1\} & \forall i \in \Vc; j\in \Vc; k \in \Kc \label{eq:mdvrp_const11} 
\end{align}

The objective function \eqref{eq:mdvrp_obj} minimizes the total routing cost for vehicles across all depots. 
Constraints \eqref{eq:mdvrp_const1} ensures that each customer is visited exactly once in the optimal solution. 
Flow conservation is enforced using constraints \eqref{eq:mdvrp_const3}, while constraints \eqref{eq:mdvrp_const4} ensure that the total demand fulfilled by a vehicle does not exceed its capacity.
Additionally, constraints \eqref{eq:mdvrp_const6} indicate that each vehicle belonging to a depot is utilized for routing at most once. 
Subtour elimination constraints are presented in \eqref{eq:mdvrp_const7}.
Constraints \eqref{eq:mdvrp_const8} prevent direct tours between depots, and constraints \eqref{eq:mdvrp_const9} distinctly partition the vehicles between depots. 
Our MDVRP formulation captures the fundamental aspects of vehicle routing with multiple depots and excludes additional constraints such as service times, depot capacities, and vehicle opening costs.

Next, we examine the decomposition of the MDVRP model in \eqref{eq:mdvrp_obj}--\eqref{eq:mdvrp_const11} into subproblems.
As evident from the formulation, vehicles are distinctly categorized based on their assigned depots, and inter-depot routes are strictly prohibited. 
Thus, once customers are assigned to the depots, the problem decomposes into $|\mathcal{D}|$ distinct CVRP subproblems.
Let $\Cc^P = \{\Cc_1, \Cc_2, \dots, \Cc_{|\Dc|}\}$ be a partition of $\Cc$ such that $\Cc = \Cc_1 \cup \Cc_2 \cup \cdots \cup \Cc_{|\Dc|}$ and $\Cc_i \cap \Cc_j =\emptyset$ for $1\leq i < j \leq |\Dc|$. 
Additionally, we assume that each depot serves at least one customer, i.e., $|\Cc_d|\geq 1, \forall d \in \Dc$. 
The partition $\Cc^P$ represents the distinct assignment of $|\Cc|$ customers among $|\Dc|$ depots.
We assume that the partition is ordered so that $\Cc_d$ is served by depot $d$. 

Let $\Vc_d = \{d\} \cup \Cc_d, \forall d \in \Dc$. With a partition $\Cc^P$, the original problem reduces to $|\Dc|$ CVRPs, and a subproblem $d \in \Dc$ is represented as follows:
\begin{align}
    \min\ & \sum_{i \in \Vc_d} \sum_{j \in \Vc_d} \sum_{k \in \Kc_d} c_{ij} x_{ijk} & \label{eq:cvrp_obj} \\
    \text{s.t.}\ & \sum_{i \in \Vc_d} \sum_{k \in \Kc_d} x_{ijk} = 1 & \forall j \in \Cc_d \label{eq:cvrp_const1} \\
    & \sum_{i \in \Vc_d} x_{ihk} - \sum_{j \in \Vc_d} x_{hjk} = 0 & \forall k \in \Kc_d; h \in \Vc_d \label{eq:cvrp_const3} \\
    & \sum_{i \in \Cc_d} \sum_{j \in \Vc_d} q_i x_{ijk} \leq Q_k & \forall k \in \Kc_d \label{eq:cvrp_const4} \\
    & \sum_{i \in \Cc_d} x_{idk} \leq 1 & \forall k \in \Kc_d \label{eq:cvrp_const6} \\
    & \sum_{i \in \Sc} \sum_{j \in \Sc} x_{ijk} \leq |\Sc|-1 & \forall \Sc \subseteq \Cc_d; |\Sc| \geq 2; k \in \Kc_d \label{eq:cvrp_const7} \\
    & x_{ijk} \in \{0, 1\} & \forall i \in \Vc_d; j\in \Vc_d; k \in \Kc_d \label{eq:cvrp_const11} 
\end{align}

The subproblem formulation \eqref{eq:cvrp_obj}--\eqref{eq:cvrp_const11} for depot $d$, obtained after decomposition, represents the well-known CVRP three-index flow formulation. 
By partitioning customers across different depots, the complexity of the original problem was significantly reduced due to the decrease in the number of feasible solutions. 
The quality of a decomposition strategy can be estimated based on the sum of the optimal routing costs of its subproblems. 
However, examining multiple partitions by solving all subproblems further increases computational difficulty.
In this study, we introduce an approach to finding good higher-level decisions using a simple heuristic and machine learning. 
Note that a random partition $\Cc^P$ may violate the vehicle capacity and fleet number constraints for one or more depots. 
Therefore, efforts must be made to carefully avoid any such infeasibility when possible. 
For instance, an undesirable partition occurs when the total demand of customers assigned to a depot surpasses the total vehicle capacity available at that depot. 
However, detecting feasibility based on higher-level decisions can be challenging and may only become apparent during subproblem resolution. 
A similar decomposition methodology can be constructed for the CLRP, which incorporates additional factors such as depot capacities, depot costs, and vehicle opening costs to evaluate the decomposition, along with the total cost of subproblems.

\subsection{General Discussion}

In general, decomposing an HVRP into CVRPs can be performed by clustering customers and/or assigning values to the higher-level decision variables. 
For example, in the CLRP, additional higher-level decisions are required to determine which depots to activate for service.
Many VRP variants, such as the orienteering Problem, the prize-collecting VRP, the periodic VRP, the split delivery VRP, the CLRP, and others, can be addressed through the decomposition method described above.
Suppose $\mathbf{\hat{x}_d}$ represent an optimal solution to a CVRP subproblem $d$ such that $\mathbf{\hat{x}} = (\mathbf{\hat{x}_d})_{d \in \Dc}$ denotes all the routes for the main problem, and let $P(\cdot)$ be a function that calculates the optimal routing cost.
$P(\mathbf{\hat{x}}) = \sum_{d \in \Dc} P(\mathbf{\hat{x}_d}; \mathbf{\hat{y}})$ provides the total optimal routing cost for the original HVRP, contingent upon higher-level decisions $\mathbf{\hat{y}}$ that determine customer partitioning and vehicle assignments. 
$P(\mathbf{\hat{x}})$ serves as a metric for evaluating the decomposition in conjunction with the costs related to the higher-level decisions $\mathbf{\hat{y}}$, denoted as $R(\mathbf{\hat{y}}, \mathbf{\hat{x}})$. 

Although this is a direct decomposition approach, there has been limited research on related methodologies due to the challenge of evaluating multiple decomposition strategies by addressing several subproblems.
Our solution to this issue involves training a supervised learning network to predict the optimal routing cost of a CVRP based on the input graph structure.
Since the neural network is trained only to predict the near-optimal cost rather than finding the actual routing solution, we benefit from significant computational advantages.
Consequently, to evaluate higher-level decisions, we need $P(\mathbf{x})$ and do not directly require $\mathbf{x} \in \mathbf{X}$.
The next section discusses the neural network architecture used to approximate the optimal cost of a subproblem.

\section{Learning to predict CVRP Costs} \label{sec:nn}

For a formal definition of our CVRP training instances, consider the complete and undirected graph $\Gc = (\Vc, \Ec)$ defined in Section \ref{sec:hvrp}, with $|\Vc| = N+1$ and the cardinality of depot set $|\Dc| = 1$.
Each customer $i$ has a positive demand denoted as $q_i, i = 1, 2, \dots, N$. The objective of CVRP is to find the minimum total cost for all routes where each route is traversed by one of the fixed number of homogeneous vehicles, each with a capacity of $Q$, satisfying the following conditions:
\begin{itemize}
    \item Each route must start and end at the depot.
    \item Each customer must belong to exactly one route.
    \item The total demand of customers assigned to a route must not exceed the vehicle capacity $Q$.
\end{itemize}

\begin{figure}
    \centering
    \includegraphics[width=\textwidth]{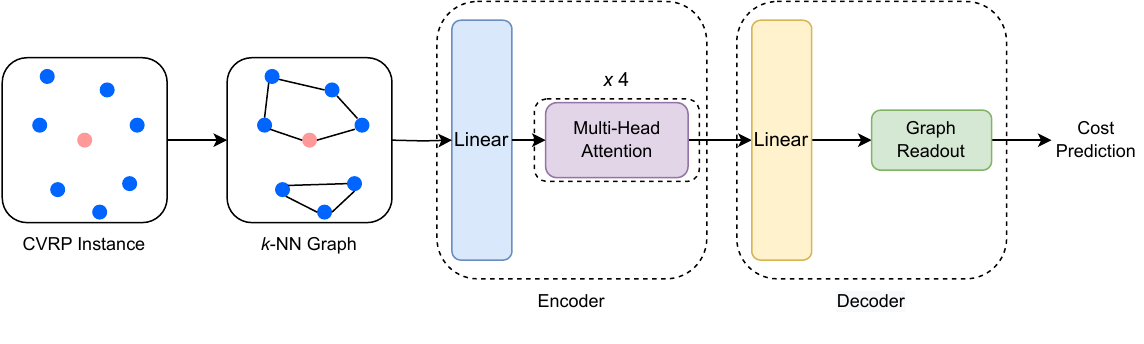}
    \caption{NN architecture}
    \label{fig:neuralnet}
\end{figure}

We assume homogeneity in the vehicle capacities and do not impose an upper bound on the number of vehicles.
In the following section, we describe our cost prediction approach for CVRPs using a graph neural network (GNN). 
Using supervised learning, GNN is trained to predict the optimal CVRP cost estimated by the HGS-CVRP solver. 
An overview of the prediction model is presented in Figure \ref{fig:neuralnet}.

\subsection{Cost Prediction using Graph Neural Network}

\paragraph*{KNN Graph Construction} 
We transform an input CVRP instance $\Gc$ into a graph-structured data $\Gc_{\text{knn}}$ to better represent the node-level and edge-level features for effectively training the GNN. 
Here, $\Gc_{\text{knn}}$ is a graph transformation of $\Gc$ constructed using the k-nearest neighbor (KNN) method.
During the KNN graph construction, each node is connected to its $k$ nearest neighbors based on the Euclidean distances. 
The node features $\mathbf{U}$ consists of two components: normalized node coordinates and normalized demand. 
The normalized node coordinates of $\Gc_{\text{knn}}$ are computed by translating the node coordinates to ensure they are positive. 
This is achieved by adding the absolute value of the minimum coordinate values and then dividing them by the maximum coordinate value.
The normalized demand is defined as the ratio of the demand $q_i$ to the vehicle capacity $Q$.

\paragraph*{Graph Neural Network Architecture} 
The cost prediction GNN $f_{\boldsymbol{\theta}}$ with parameters ${\boldsymbol{\theta}}$ predicts the optimal cost of $\Gc$ as $\hat{v}$. 
Our GNN architecture comprises the encoder, which consists of a stack of self-attention blocks \citep{vaswani2017attention}, and the decoder, a linear layer followed by the graph readout module. 
$f_{{\boldsymbol{\theta}}}$ predicts the cost as follows:
\[
\begin{aligned}
	&\Gc_{\text{knn}} \xleftarrow[\text{construction}]{\text{KNN graph}} \Gc  \\
    &\hat v = f_{{\boldsymbol{\theta}}}(\Gc_{\text{knn}}) \coloneqq \texttt{Decoder}(\texttt{Encoder}(\Gc_{\text{knn}})),
\end{aligned}
\]
where $\texttt{Encoder}(\cdot)$ and $\texttt{Decoder}(\cdot)$ are described in the following paragraphs.

The GNN encoder captures meaningful information about the input graph data in the form of a latent representation.
The $\texttt{Encoder}(\cdot)$ operator transforms $\Gc_{\text{knn}}$ into the latent vectors $\mathbf{U'} \in \mathbb{R}^{(N+1) \times h}$, where $h$ denotes the hidden node feature dimension.
This transformation is achieved through a series of operations.
First, a linear embedding layer is applied to match the input node feature dimension $\tilde{d}$ to $h$. Then, multiple self-attention blocks are employed to learn the attention scores with respect to the extracted node features. 
Each self-attention block consists of two steps: $\texttt{MHA}(\cdot)$ and $\texttt{FeedForward}(\cdot)$.
The Multi-Head Attention (MHA) mechanism divides the input into multiple heads and computes the attention for each head simultaneously.
Each head may prioritize different aspects of the input data, leading to focused learning and improved accuracy. 

The use of MHA in the graph encoder exhibits multiple benefits over other architectures such as Multilayer Perceptron (MLP).
Firstly, MHA handles an arbitrary number of input nodes within a single model. 
In contrast, traditional networks such as MLP are incapable of handling variable-length data and this limits the generalizability of learning to predict a general VRP instance.
Another significant advantage of MHA is its representational capability for complex graph data. 
By dynamically adjusting the strength of relationships among nodes based on their features, MHA can disregard the correlations between the nodes that do not impact the CVRP cost prediction. 
On the contrary, the prediction accuracy of conventional GNN architectures like Graph Convolutional Networks and Graph Attention Networks relies on the design of the input graph structure. 
The dynamic edge weight allocation of MHA mitigates the errors arising from suboptimal graph designs.
Furthermore, MHA offers higher computational efficiency with matrix multiplication on modern computing frameworks compared to the message-passing frameworks in traditional GNNs, leading to faster training and inference.

If $\text{head}_i$ denotes the attention applied to $\mathbf{U}$ for $i \in \{1,2,\dots,H\}$, and for a learnable parameter $\mathbf{W^O} \in \mathbb{R}^{H \times h}$, the $\texttt{MHA}(\cdot)$ step is defined as follows:
\[
\begin{aligned}
	\text{head}_i &= \texttt{Attention}(\mathbf{U}) \\
	\texttt{MHA}(\mathbf{U}) &= \text{Concat}(\text{head}_1, \dots, \text{head}_H) \cdot \mathbf{W^O}
\end{aligned}
\]
where the $\texttt{Attention}(\cdot)$ transformation is applied to each node $u_i$ in $\mathbf{U}$ to compute the updated node feature $u'_i$, given by
\begin{equation*}
u'_i = \sum_{j \in \mathcal{N}(i)} \underset{j \in \mathcal{N}(i)}{\text{{softmax}}} \left(\frac{\mathbf{W^Q}u_i \cdot \mathbf{W^K}u_j}{\sqrt{h}}\right) \cdot \mathbf{W^V}u_j
\end{equation*}
where $\mathbf{W^Q}$, $\mathbf{W^K}$, and $\mathbf{W^V} \in \mathbb{R}^{h \times h}$ are the learnable parameters, $\mathcal{N}(i)$ represents the set of neighbors of node $i$, and $\text{{softmax}}$ denotes the softmax function applied over $j \in \mathcal{N}(i)$. 
After the Multi-Head Attention (MHA) step, the updated node features $\mathbf{U'}$ are obtained. 
The attention mechanism is applied to the neighbors of each node, and the attention weights are computed based on the similarity between the node features. This approach allows the encoder to capture the local relationships between the nodes.

\noindent After accounting for the local relationships between the nodes, $\texttt{Encoder}(\cdot)$ applies $\texttt{FeedForward}(\cdot)$ to process nodes independently. The $\texttt{FeedForward}(\cdot)$ operator is a multilayer perceptron model applied to each node feature. 
The updated node features $\mathbf{U'}$ are computed using the normalization layers $\texttt{LayerNorm}(\cdot)$ \citep{ba2016layer} and residual connections \citep{he2016deep}, which are applied to enhance the trainability of the deep models.
To summarize, $\texttt{Encoder}(\cdot)$ consists of the following sequential steps:
\[
\begin{aligned}
\mathbf{U'} &= \texttt{{LayerNorm}}(\mathbf{U} + \texttt{{MHA}}(\mathbf{U})) \\
\mathbf{U'} &\leftarrow \texttt{{LayerNorm}}(\mathbf{U'} + \texttt{{FeedForward}}(\mathbf{U'}))
\end{aligned}
\]

The $\texttt{Decoder}(\cdot)$ operator subsequently transforms the latent vectors $\mathbf{U'}$ into the predicted optimal cost $\hat v$. This transformation involves a linear layer to project the latent vectors into a lower-dimensional space, followed by a graph readout module to aggregate the graph information. $\texttt{Decoder}(\cdot)$ is defined as follows:
\begin{equation*}
    \hat v = \frac{1}{|\mathcal{V}|} \bigg( \mathbf{1}_{|\mathcal{V}|} (\mathbf{U'}\mathbf{W} + \mathbf{b} ) \bigg), 
\end{equation*}
where $\Vc$ represents the set of nodes in $\Gc_{\text{knn}}$, $\mathbf{1}_{|\mathcal{V}|} = (1, 1, \dots, 1) \in \mathbb{R}^{1 \times |\mathcal{V}|}$, and $\mathbf{W} \in \mathbb{R}^{h \times 1}$ and $\mathbf{b} \in \mathbb{R}^{|\mathcal{V}| \times 1}$ are the scalar projection parameters.
The graph readout module calculates a weighted average of the transformed latent vectors to produce a scalar output. This approach allows the decoder to capture the global relationships between the nodes.

\subsection{Training}
Existing CVRP benchmark datasets are insufficient in quantity to train this model, and thus, based on the problem requirements, CVRP instances need to be generated and solved to optimality (or near-optimality) to form the training data. 
An effective and widely used heuristic method for this problem is the Hybrid Genetic Search (HGS) for CVRP by \citet{vidal2022hybrid}, which has an open-source implementation.
HGS-CVRP solver remains a top-performing metaheuristic in terms of methodological simplicity, solution quality, and convergence speed. 
We utilize a key feature of this solver by setting a time limit to compute the near-optimal routing costs for the train data, especially for larger instances.

Due to input data normalization, the final cost from average pooling needs to be rescaled by a corresponding factor to obtain the final cost prediction of the CVRP instance. 
We use the simple yet effective mean squared error as our loss function \eqref{eq:loss_function}.
\begin{equation}
    L(\boldsymbol{\theta}, \mathbf{U}) = (f_{\boldsymbol{\theta}}(\Gc_{\text{knn}}) - \texttt{Solver}_\text{CVRP}(\Gc))^2 \label{eq:loss_function}
\end{equation}

\section{Solution Methodology for MDVRP} \label{sec:mdvrp}

In this section, we propose our solution methodology to solve large-scale MDVRPs that involve the decomposition of the original problem into different CVRP subproblems.
This approach aims to determine the optimal decomposition of the MDVRP into subsequent subproblems, where each CVRP instance corresponds to a depot of the main problem.
Using an effective decomposition strategy, the desired CVRP subproblems can be obtained and solved using any existing method or solver.
As CVRPs have been extensively studied, we do not attempt to develop a novel method for solving them.
Instead, we utilize a well-established and validated heuristic solver to arrive at the final solution.
An evaluation of all possible decomposition strategies could be computationally intractable. 
Therefore, we first select a set of random customer assignment strategies and improve them iteratively with a simplified genetic algorithm.
Note that evaluating each MDVRP decomposition in the GA with an existing VRP solver is computationally expensive, and it defeats our goal of solving MDVRPs more effectively than the existing methods in the literature.
With a supervised deep learning model, we predict the optimal routing cost of a CVRP instance almost instantly without actually solving it.

\begin{figure}
    \centering
    \includegraphics[scale=1.0]{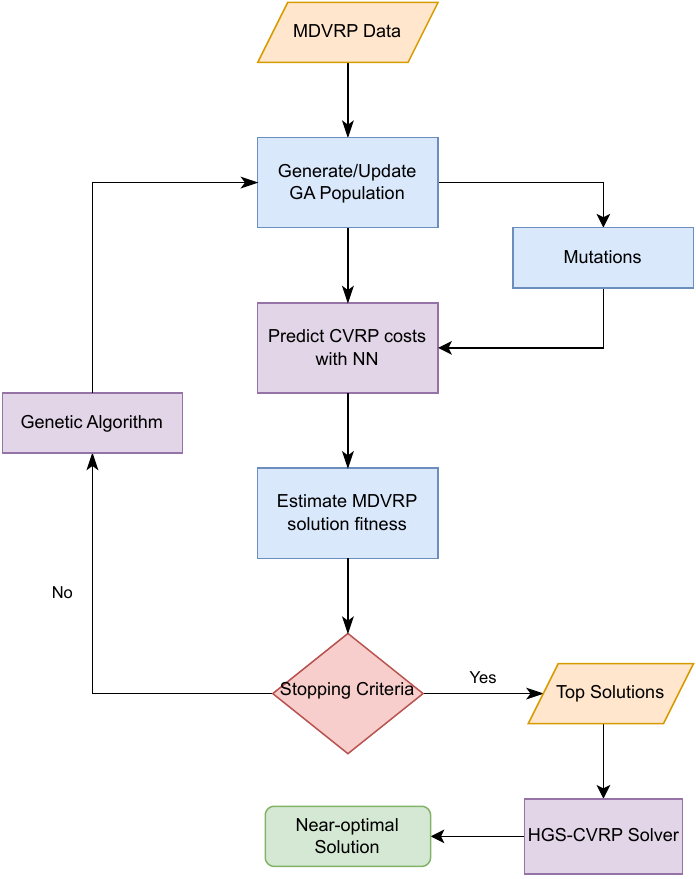}
    \caption{An outline of $\text{GANCP}^+$}
    \label{fig:outline}
\end{figure}

An outline of our methodology is illustrated in Figure \ref{fig:outline}. 
The following sections will discuss the main components of our $\text{GANCP}^+$ heuristic, such as the genetic algorithm, fitness estimation, and obtaining a final routing solution. 

\subsection{Genetic Algorithm for MDVRP decomposition}

\begin{figure}
    \centering
    \begin{tikzpicture}
        \matrix [matrix of nodes] %
        {
           $i$: & 1 & 2 & 3 & 4 & 5 & 6 & 7 & 8 \\
            & |[draw]| 2 & |[draw]| 3 & |[draw]| 1 & |[draw]| 1 & |[draw]| 3 & |[draw]| 2 & |[draw]| 1 & |[draw]| 3 \\
        };
    \end{tikzpicture}
    \caption{Chromosome representation of customer allocations to depots $\{1,2,3\}$}
    \label{fig:chromosome}
\end{figure}

As an evolutionary approach, GAs represent possible solutions as a set of chromosomes that evolve towards optimality through a process of parent selection, crossover, and mutation over multiple generations. 
To apply GA, each chromosome is encoded as a string of data. 
We represent each individual as an $N$-length chromosome of depot allocations that corresponds to customers $i \in \{1,2,\dots, N\}$, as depicted in Figure \ref{fig:chromosome}. 
The customer indices with the same depot allocation $d$ are selected to form the CVRP with depot $d$.

Algorithm \ref{alg:ga_outline} gives a comprehensive overview of our method, and it begins with the generation of an initial population that consists of multiple MDVRP decompositions. 
Random assignments of customers to depots generate a diverse initial population, and it ensures that the population starts at a high level of diversity, leading to a larger solution search space over time. 
Besides this rule, we also initialize two or three potential solutions based on the input graph structure. 
Our study found that a combination of targeted and random assignments in the initial population results in faster algorithm convergence.
One effective targeted assignment strategy is the nearest depot assignment (NDA), where the customers are allocated to their nearest depot. 
The second strategy is to allocate each customer to the closest neighbor's nearest depot.
These initial assignment rules have been discussed in \cite{de2016cooperative}.
Additionally, we initialize a chromosome where customers are allocated to their second nearest depot for problems with more than two depots.
Targeted assignments serve as an effective initial solution to problems involving uniform customer distribution in a region.
Metaheuristics can improve such initial solutions and achieve near-optimal results.

During the solution search, the genetic algorithm may encounter infeasible candidates.
While all the infeasible candidates cannot be determined during the cost prediction phase of our heuristic, we can detect candidates with a high degree of infeasibility.
For example, a VRP is evidently infeasible if the total demand exceeds the total vehicle capacity.
Such individuals are repaired by selecting a random customer from the infeasible CVRP and reassigning them to a depot that can fulfill the corresponding demand, as detailed in Algorithm \ref{alg:repair_infeasibles}.

\begin{algorithm}
    \caption{$\text{GANCP}^+$} \label{alg:ga_outline}
    \begin{algorithmic}[1]
        \State ${\boldsymbol{\Omega}} \gets \emptyset$ \Comment{Feasible and Infeasible Population}
        \State ${\boldsymbol{\Omega}} = \texttt{initial\_population()}$ 
        \State $\texttt{repair\_infeasibles}({\boldsymbol{\Omega}})$
        \State ${\boldsymbol{\hat S}} = \texttt{fitness\_scores}(f_{{\boldsymbol{\theta}}}, {\boldsymbol{\Omega}})$
        \While{number of generations $<N_{\textsc{G}}$}
        \State ${\boldsymbol{\Omega}}_\textsc{sub} = \texttt{sub\_population}({\boldsymbol{\Omega}}, {\boldsymbol{\hat S}}, P_\textsc{l})$ \Comment{Subpopulation}
        \State $\texttt{repair\_infeasibles}({\boldsymbol{\Omega}}_\textsc{sub})$
        \State ${\boldsymbol{\hat S}} = \texttt{fitness\_scores}(f_{{\boldsymbol{\theta}}}, {\boldsymbol{\Omega}}_\textsc{sub})$
        \State ${\boldsymbol{\Omega}}_\textsc{sub} \gets \texttt{mutations}({\boldsymbol{\Omega}}_\textsc{sub}, {\boldsymbol{\hat S}}, \textsc{elite})$
        \State ${\boldsymbol{\hat S}} = \texttt{fitness\_scores}(f_{{\boldsymbol{\theta}}}, {\boldsymbol{\Omega}}_\textsc{sub})$
        \State ${\boldsymbol{\Omega}} = \texttt{top\_candidates}({\boldsymbol{\Omega}}_\textsc{sub}, {\boldsymbol{\hat S}}, {\boldsymbol{\Omega}}, P_\textsc{H})$ 
        \EndWhile
        \State $\texttt{evaluate\_best\_candidates}({\boldsymbol{\Omega}}, \texttt{CVRP\_Solver})$
        \State Return $\texttt{best\_solutions}()$
    \end{algorithmic}
\end{algorithm}

\begin{algorithm}
    \caption{Repair Infeasible Assignments} \label{alg:repair_infeasibles}
    \begin{algorithmic}[1]
        \For {$\omega$ in ${\boldsymbol{\Omega}}$} \Comment{Infeasible candidate}
            \State $r \gets \texttt{rand}(0,1)$
            \If {$r<p_\textsc{repair}$}
                \While{$\omega$ is infeasible assignment} 
                    \State Find $\texttt{VRP}^d_{\omega, \textsc{INF}}$ with $\sum_{i \in \Cc_d} q_i > \sum_{k \in \Kc_d} Q_k$ for $d \in \Dc$ \Comment{Infeasible VRP}
                    \State Randomly remove customer $i$ in $\texttt{VRP}^d_{\omega, \textsc{INF}}$
                    \State Find $\texttt{VRP}^d_{\omega, \textsc{INF}}$ with sufficient remaining capacity. \Comment{Feasible VRP}
                    \State $\texttt{VRP}^d_{\omega, \textsc{INF}} \gets \texttt{VRP}^d_{\omega, \textsc{INF}} \cup i$
                \EndWhile
            \EndIf
        \EndFor
        \State Return ${\boldsymbol{\Omega}}$
    \end{algorithmic}
\end{algorithm}

The next generation of the population is propagated using multiple iterations. At each iteration, two distinct parents are selected from the current solution pool based on fitness scores using a $K$ tournament selection method. 
As larger tournament sets select better chromosomes but require more computation and may result in premature convergence, we use the binary tournament selection procedure.
To generate children, uniform crossover operations are performed over the selected parents by randomly selecting offspring genes from the corresponding genes of the parents. Our simple chromosome representation requires minimal information about the routing solution, so more sophisticated crossover operations are not necessary.
The population size is maintained in the range $[P_L, P_H]$. Mutations are performed once crossover operations have generated $P_L$ chromosomes for the next generation.

Mutations include the random reset of a customer's depot called \textit{FLIP}, and the \textit{SWAP} operation in which two selected customers' depots are interchanged. 
Based on their fitness score, chromosomes in the first tertile are selected for mutation. 
Approximately $5\%$ of all customers undergo a mutation, resulting in either a \textit{FLIP} or \textit{SWAP} operation with a probability of $p_m$ or $1-p_m$, respectively. 
Additional targeted mutations are also carried out over $5\%$ of the child population using a binary tournament selection process. 
As part of targeted mutations, $10\%$ of the customers are randomly selected, and the depot allocations of these individuals are copied from one of the elite individuals within the initial population of targeted assignments.

\subsection{NN Estimation and Fitness function} 
A candidate solution generated by the GA should be evaluated using a measure of fitness to guide the population toward the direction of minimum cost. 
In the MDVRP, the solution cost is simply the sum of the optimal routing costs for its CVRP subproblems. 
These costs are predicted using the NN prediction model trained with parameters ${\boldsymbol{\theta}}$. Although our objective is to minimize the routing cost of the main problem, the application of the MDVRP cost directly as a measure of fitness may lead to premature convergence. 
With a score for the diversity of a solution in the population, in addition to the predicted cost, which measures solution quality, the GA expands its search space and often overcomes local optima. 
Hence, the fitness of an individual can be determined using the predicted MDVRP cost \eqref{eq:mdvrp_cost} and the diversity factor \eqref{eq:diversity}. 
The diversity factor of a chromosome is measured as the mean Hamming distance of the individual from the rest of the population, as shown in \eqref{eq:diversity}.
The fitness score $\text{fit}(I)$ \eqref{eq:fitnessfunction} for a chromosome $I$ is calculated as the weighted sum of normalized solution cost $\tilde{C}$, normalized diversity factor $\tilde{\delta}$, and a penalty for the extent of approximate vehicle capacity violations. 
Note that $l_d$, $m_d$, and $Q$ represent the total demand, the number of available vehicles, and homogeneous vehicle capacity at the depot $d$, respectively. 
\begin{align}
    \delta(I) &= \frac{1}{N |{\boldsymbol{\Omega}}|}  \sum_{\substack{I' \in {\boldsymbol{\Omega}} \\ I' \neq I}} \sum_{i = 1}^N \mathbbm{1} (I(i)\neq I'(i)) \label{eq:diversity} \\
    C(I) &= \sum_{d \in \Dc} f_{{\boldsymbol{\theta}}}(\mathsf{VRP}(\Vc_d)) \label{eq:mdvrp_cost} \\
    \text{fit}(I) &= w_1 \tilde{C}(I) - w_2 \tilde{\delta}(I) + 
    w_3 \tilde{C}(I) \sum_{d \in \Dc} (l_d-m_dQ)^+
    \label{eq:fitnessfunction}
\end{align}

Newly generated chromosomes from crossovers and mutations are evaluated in batches to find the MDVRP cost predictions, and duplicate solutions are eliminated before the batch processing.
When the CVRP subproblems from multiple chromosomes are processed together with the NN model, the computational time reduces significantly. 
Following this, fitness scores are calculated for the new generation. 
According to the fitness scores, a group of $P_H$ individuals are selected and represent the next generation. 
To prevent wide divergences from our objective, the top $1\%$ of individuals from each generation is preserved for the subsequent generation. 
This process is repeated until there is no improvement over multiple generations or up to the desired number of generations.

\subsection{Determining the final solution by HGS-CVRP}
Once $\text{GANCP}$ reaches the stopping criteria, producing potential solutions with cost predictions, the next step is to determine the final routing solution. 
Since solution diversification is no longer crucial, we rank the population and the top individuals from previous generations based on the predicted MDVRP cost. 
At this stage, $\text{GANCP}$ generates good-quality decompositions, yet the routing solutions to the CVRP subproblems remain unknown.
To obtain the actual MDVRP solution cost and the corresponding vehicle routes, we utilize the HGS-CVRP solver \citep{vidal2022hybrid} with a manual time limit for the solver based on an approximate input subproblem size. 
This approach facilitates obtaining near-optimal solutions for subproblems of varying sizes within reasonable timeframes through the utilization of the HGS-CVRP solver.
Furthermore, as machine learning predictions may contain errors, we evaluate the top $k$ individuals ranked according to our genetic algorithm, using the HGS-CVRP solver, to obtain a better quality solution for the MDVRP. 
Our prediction-based heuristic, $\text{GANCP}$, with the HGS-CVRP solver at the final stage, is referred to as $\text{GANCP}^+$.

\section{Computational Study for MDVRP} \label{sec:comp_exp1}

In this section, we discuss the training procedure of the NN model for CVRPs, followed by MDVRP large-scale experiments and tests for the out-of-range and out-of-distribution robustness of the proposed method. 
All the computational experiments have been performed in a computer with 11th Gen Intel(R) Core(TM) i9-11900H 2.50GHz processor, 32GB RAM, 8 CPU cores with 2 logical processors each, and NVIDIA GeForce RTX 3080 Laptop GPU with 8 GB memory, running Windows 11. 
We implemented the NN model with the same hyperparameters reported in \citet{li2021learning} and tuned the number of layers using grid search.
The NN architecture is implemented and trained in Python 3.9, while our proposed algorithm is implemented in Julia 1.7.2 with function calls to Python for cost predictions using a Julia package $\text{PyCall}$. 
$\text{GANCP}^+$ is adapted to minimize the number of Python calls by employing batch estimation of VRP costs.
Our code is publicly available at \url{https://github.com/abhaysobhanan/GANCP}.

\subsection{NN Training}
In this study, we explore the effectiveness of different CVRP train data generation strategies, referred to as phases, in ranking the decomposed MDVRP subproblems.
Each training phase consists of $128,000$ VRP instances, each with a random number of customer locations $N \in [50,500]$. 
Random samples are taken from the generated data in a ratio of $80:20$ for both training and validation. 
For the evaluation of the trained model, a test set consisting of $1000$ random VRP instances is generated separately. 
To determine the optimal or near-optimal routing cost for each VRP instance used in the supervised learning, the HGS-CVRP solver is executed with a time limit of 5 seconds.
It is recommended to use parallel processing to speed up the process of finding CVRP costs using the HGS-CVRP solver.
We utilize established CVRP cost approximation expressions from \citet{robust1990implementing} and \citet{figliozzi2008planning} to assess the prediction accuracy of our GNN architecture, presented in \eqref{eq:dist_daganzo} and \eqref{eq:dist_approx_2008}.
Due to the existing literature studies \citep{xu2020neural, cappart2023combinatorial} indicating that GNNs outperform other neural network architectures, particularly when handling variable-length graph data of large scale, our comparison focuses on the analytical expressions. 

\begin{figure}
    \centering
    \includegraphics[scale=0.8]{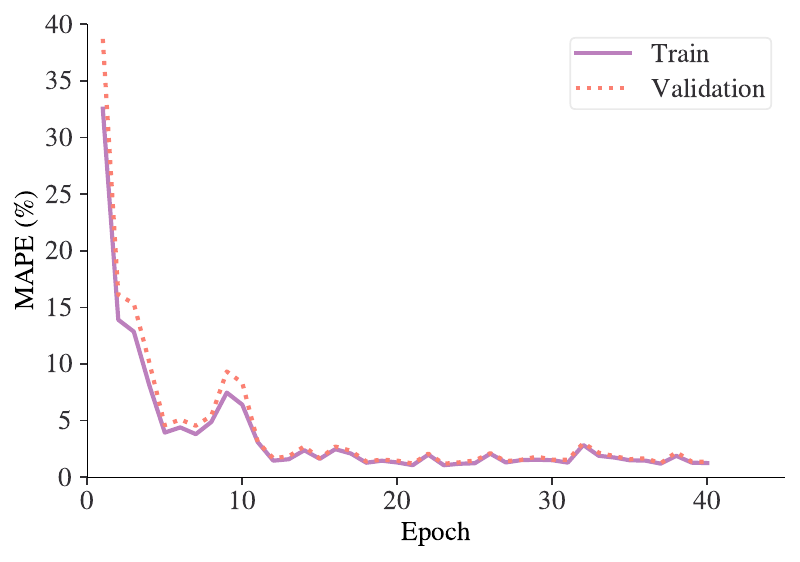}
    \caption{Training progress of the supervised learning NN model during Phase \RNum{1}}
    \label{fig:training_curve}
\end{figure}

\begin{figure}%
    \centering
    \subfloat[\centering Near-optimal vs Predicted Costs]
    {
        {\includegraphics[width=8cm]{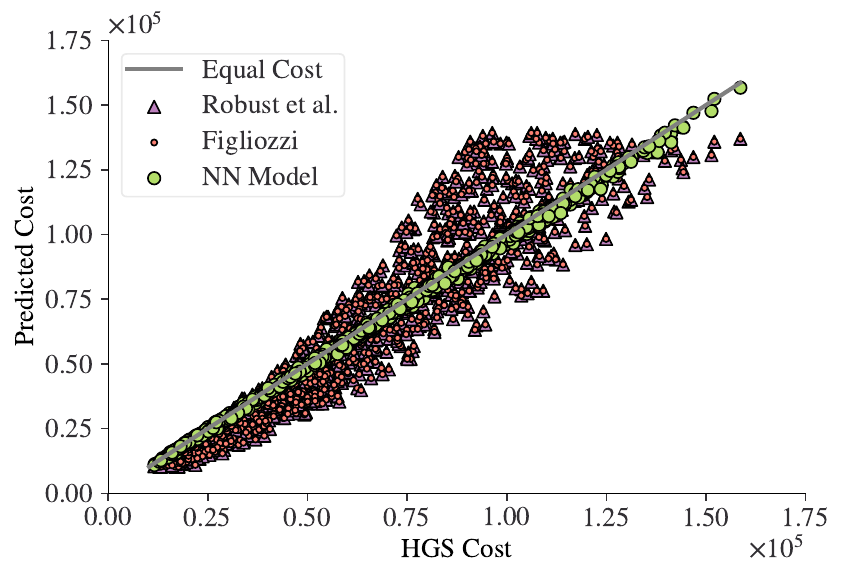}}
    }
    \subfloat[\centering Prediction error for different problem sizes]
    {
        {\includegraphics[width=8cm]{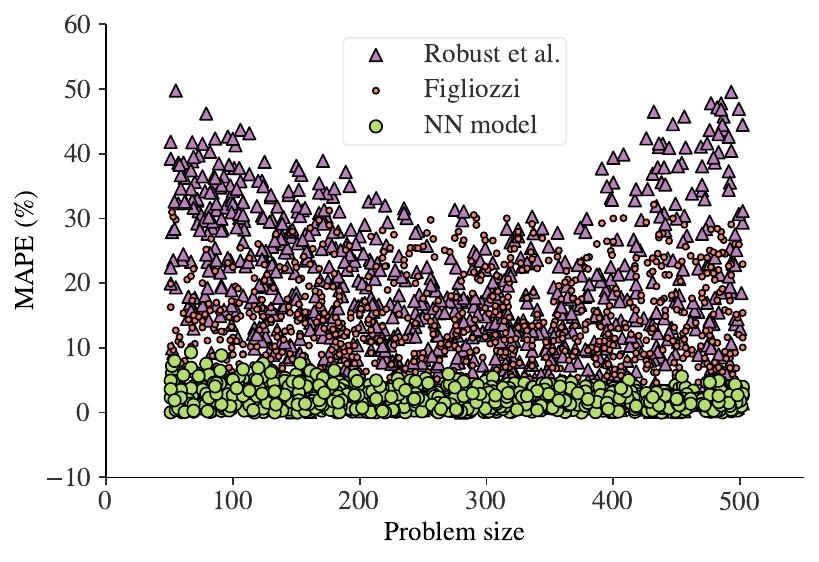}}%
    }
    \caption{\centering Phase \RNum{1} test results of NN trained with random VRP instances where $N \in [50,500]$}
    \label{fig:NN_TestResults}%
\end{figure}

A common approach for training the proposed NN on CVRP data is to randomly distribute customers across a grid. 
However, we demonstrate that this method is inadequate for accurately predicting the subproblems arising from optimal MDVRP decompositions. 
To address this, we conduct three distinct training sessions on the proposed NN, each utilizing a different data generation technique, referred to as a phase.
In Phase \RNum{1}, the VRP instances are generated following the rules of \citet{uchoa2017new} with randomly placed depots and customers, where coordinates are integers from a grid of size $[0,1000] \times [0,1000]$ and demands are integers in the range $[1,100]$. 
Phase \RNum{1} follows the common approach discussed above and is effective for estimating the cost of randomly allocating customers to depots.
The learning curves for both the training and validation data are illustrated in Figure \ref{fig:training_curve}.
While this trained model provides good cost estimates for the initial MDVRP solutions, the prediction error increases significantly as the GA propagates toward optimality. 
Figure \ref{fig:NN_TestResults} depicts the testing accuracy of Phase-\RNum{1} with an average mean absolute percentage error (MAPE) of $1.06\%$, and shows that our NN model outperforms \eqref{eq:dist_daganzo} and \eqref{eq:dist_approx_2008} significantly. 
We choose $C=6.52$ in \eqref{eq:dist_daganzo} to minimize the average error across the test data. 
Similarly, the regression parameters in \eqref{eq:dist_approx_2008} are obtained using the same train data in Phase \RNum{1}. 

In Phase \RNum{2},  the VRP subproblem instances are generated from random MDVRP instances to incorporate the near-optimal cases seen during our methodology. 
$80\%$ of the training data is generated using the two targeted assignment rules, out of which in $70\%$ of the instances, exceptions are made to the targeted assignments by randomly reallocating up to $10\%$ of the customers.
The remaining $20\%$ of the instances follow Phase \RNum{1} data generation.
Phase \RNum{2} involves targeted train data based on a partial understanding of our real data distribution. 
As a result of comparing multiple assignments, MDVRP solutions are better represented, and cost prediction errors are reduced towards optimal decompositions.

In the final training phase, i.e., Phase \RNum{3}, we generate train/validation data in four steps using random MDVRP instances in conjunction with $\text{GANCP}$. 
In the first step, the NN prediction model is initialized with randomized parameters ${\boldsymbol{\theta}}$. 
For an input CVRP, this untrained model is incapable of a fair cost prediction and provides any real number as the output based on ${\boldsymbol{\theta}}$. 
Therefore, the final solutions from $\text{GANCP}$ are random assignments. 
Multiple MDVRP instances are utilized to generate CVRP subproblems.
After each step, the VRP subproblems generated based on the existing ${\boldsymbol{\theta}}$ are evaluated with the HGS-CVRP solver.
This serves as the train data for the NN model, and the parameter values are updated.
In the next step, the improved NN model parameters are used in $\text{GANCP}$ with random MDVRP instances to generate the new VRP decompositions. 
This process of NN parameter estimation using partial train data is repeated until the NN model is trained in the fourth step.
For each MDVRP evaluation, the top $2$ solutions using $\text{GANCP}$ are selected to produce the corresponding CVRP subproblems. 
Each step contributes to the generation of $1/4$-th of the total train/validation data.
At each step, the generated data is used for training in combination with data from previous steps, if applicable. 
This enhances the capability of the NN model to evaluate the CVRP instances arising from various assignment patterns observed during our heuristic. 

\input{tables/hvrp_table_cvrp.tex}

\subsection{Test instances}
For evaluation of our $\text{GANCP}^+$ methodology, $150$ MDVRP Test (T) instances are generated in the range $N \in [100,1500]$ and $D \in [2,10]$. The number of depots for each problem is chosen so that the approximate CVRP subproblem size falls within the range of trained data, i.e., $\frac{N}{D} \in [50,500]$. 
Instances are generated based on random customer and depot positioning in a grid of size $[0,1000] \times [0,1000]$, while demands are integers in the range of $[1,100]$. 
The NN models from each phase undergo testing using 20,000 random CVRP instances, and the results are summarized in Table~\ref{tab:cvrp_avg} for different problem sizes.
The performance gap $g_H = \frac{\text{Predicted Cost} - \text{HGS Cost}}{\text{HGS Cost}}\times 100$ is calculated in comparison to the HGS-CVRP solver, and the absolute values $|g_H|$ are reported to consider both overestimation and underestimation. 

Table~\ref{tab:ModelAccuracies} presents the performance accuracy of the NN model trained in different phases, both on random CVRP instances and when using our heuristic to obtain MDVRP solutions. 
We summarize the results for both the average and best of 10 experiments.
The average performance gap (\%) for random CVRP instances is calculated as the absolute percentage deviation from the cost obtained by the HGS-CVRP solver, while the performance gap for T instances is calculated against a benchmark solver.
A negative gap for T instances indicates that  $\text{GANCP}^+$ obtains a better quality solution compared to the benchmark solver.
The details regarding the experiments on T instances and the benchmark solver will be discussed subsequently.
While Phase \RNum{1} demonstrates strong performance for random CVRP data, the training falls short in incorporating VRPs that arise from the near-optimal MDVRP decompositions. 
Given that Phase \RNum{3} excels for our MDVRP heuristic, the remaining experiments will focus on the NN model formed using the Phase \RNum{3} parameters, unless stated otherwise.
Note that in the event of limited training time availability, Phase \RNum{2} can be used to obtain satisfactory results.

\begin{table}
    \footnotesize
    \centering
    \caption{NN model accuracy for different training phases}
    \begin{tabular}{ l r r r }
    \toprule
    \multicolumn{1}{c}{\multirow{3}{*}{Train Data}} & \multicolumn{3}{c}{Average Performance Gap (\%) } \\
    \cmidrule(lr){2-4}
    & \multicolumn{1}{c}{\multirow{2}{*}{CVRP}} & \multicolumn{2}{c}{T Instances} \\
    \cmidrule(lr){3-4}
    & & \multicolumn{1}{c}{$\text{GANCP}^+_\text{avg}$} & \multicolumn{1}{c}{$\text{GANCP}^+_\text{best}$} \\
    \midrule
    Phase \RNum{1} & 1.06 & 0.50 & -0.49\\
    Phase \RNum{2} & 2.21 & -0.09 & -0.47\\
    Phase \RNum{3}  & 2.26 & -0.58 & -0.89\\
    \bottomrule  
    \end{tabular}
    \label{tab:ModelAccuracies}
\end{table}

\paragraph{Algorithm Settings}
The computational requirements of the HGS-CVRP solver increase as the size of the problem increases. 
The time limits for the HGS-CVRP solver based on approximate subproblem size, shown in Table \ref{tab:Hyperparameters}, are used to ensure better routing solutions for the top candidates obtained from $\text{GANCP}$.  
Additionally, the number of CVRP instances that can be processed by the NN model in a batch depends on the memory limitations of the computer. 
On the specified computer, the batch processing is executed in units as specified in Table \ref{tab:Hyperparameters} to avoid memory overflows.

\begin{table}
    \footnotesize
    \centering
    \caption{Hyperparameters used under the given system specifications}
    \begin{tabular}{c r r}
    \toprule
    \multicolumn{1}{c}{Approximate} & \multicolumn{1}{c}{NN batch size} & HGS time \\ 
    subproblem size &  & (sec) \\
    \midrule
    $N/D \in [50,100)$ & 512 & 0.1\\
    $N/D \in [100,200)$ & 256 & 1.0 \\
    $N/D \in [200,300)$  & 128 & 2.0\\
    $N/D \in [300,400)$ & 64 & 2.0\\
    $N/D \in [400,500]$ & 32 & 2.0\\
    \bottomrule   
    \end{tabular}
    \label{tab:Hyperparameters}
\end{table}

\paragraph{Benchmark Algorithm}
Table \ref{tab:cordeau} shows the performance of our method on benchmark instances in \citet{cordeau1997tabu}.
Only the instances without route-duration constraints were used in the experiment. We compare our results to the best-known solution costs in the literature.
A distinct training of the NN model using random VRP instances of small problem sizes in the range $N \in [10,50]$ was performed to evaluate the subproblems of benchmark instances. Customer and depot positioning were randomly chosen, similar to Phase \RNum{1} train data. 
As expected, $\text{GANCP}^+$ performs relatively poorly for the out-of-distribution \citet{cordeau1997tabu} instances, with a maximum gap of $4.61\%$. 
The best-known results are also compared with an open-source solver VROOM \citep{vroom} on the same machine running Ubuntu 22.04.1, and the results indicate that VROOM is a competitive solver for the MDVRP instances and yields high-quality solutions within a short time frame. 
To ensure that the comparison between the benchmark MDVRP solver and the use of our NN model on a GPU is fair, we run VROOM using 16 CPU threads on the same machine for all future experiments. 
Through the use of multiple CPU threads, we improve the benchmark solver's computational speed.

Additionally, we employ two simple decomposition rules for comparison: the nearest depot assignment (NDA) and $\text{K-Means-10}$.
NDA provides near-optimal solutions for large instances with uniform customer positioning and demand distributions.
However, this simplistic decision-based approach has two drawbacks.
Firstly, the NDA does not perform well when dealing with non-uniform data.
Secondly, the likelihood of encountering infeasible solutions increases when dealing with tighter vehicle capacities or when additional constraints such as time limits are introduced. 
For example, NDA produces an infeasible solution for the instance p07 in Table \ref{tab:cordeau}. 
To validate this infeasibility issue of NDA, we generated 105 additional test instances by introducing random perturbations to the coordinates and demands in the Cordeau instances. 
NDA produced an infeasible solution in approximately 25\% of these instances, and for the remaining instances $\text{GANCP}^+$ consistently outperformed NDA with a gap of -1.77\%.
In the $\text{K-Means-10}$ approach, customers are grouped into $D$ clusters based on randomized centroids.
After forming the clusters, each one is assigned distinctly to a depot based on the Euclidean distance between the centroids and the depots.
This process is repeated 10 times, and the best cost and total time are reported.

\input{tables/hvrp_table_cordeau.tex}

We execute $\text{GANCP}^+$ on $150$ MDVRP T instances of varying size, and a summary of the results for the same dataset in comparison to the benchmark algorithms is shown in Tables \ref{tab:T_results_summary1} and \ref{tab:T_results_summary2}. 
Table \ref{tab:T_results_summary1} displays the results obtained from NDA, $\text{K-Means-10}$, VROOM, and the best $\text{GANCP}^+$ results out of 10 experiments. 
Table \ref{tab:T_results_summary2} presents the results obtained from VROOM-$l$ and the average $\text{GANCP}^+$ result out of 10 experiments, where
VROOM-$l$ denotes the execution of VROOM with a time limit equal to the run time of $\text{GANCP}^+$. 
Appendix \ref{sec:detailed_results} contains the detailed results of the experiments. 
The performance gap of a solution with respect to NDA, $\text{K-Means-10}$, and VROOM are given by $g_N, g_K$, and $g_V$, respectively. 
For example,  we define
\[
	g_N = \frac{\text{GANCP}^+\text{ Cost} - \text{NDA Cost}}{\text{NDA Cost}} \times 100 . 
\]
A negative gap indicates a better quality $\text{GANCP}^+$ solution.
The results indicate the effectiveness of our approach in obtaining high-quality solutions at faster computational speeds. 
Among the T instances, the largest problem size considered shows that the runtime of $\text{GANCP}^+$ is 11.4 times faster than VROOM, with an observed performance gap of $-0.39\%$.
With a shorter time limit, VROOM often produces the same solutions as the full run but fails to improve the earlier solutions.
NDA yields good quality solutions for the T instances, sometimes even better than VROOM.
This is because VROOM, like most solvers, is tailored for medium-sized problems. This is evident from Table \ref{tab:cordeau} and the smallest problem size considered in Table \ref{tab:T_results_summary1}, where the NDA exhibits relatively poor performance. 
Our GANCP+ framework outperforms NDA without imposing a significant computational time requirement, as compared to VROOM, even for large instances that follow a uniform distribution and seldom exhibit infeasibility issues with the NDA.
Note that employing a random clustering approach such as $\text{K-Means-10}$ may result in low-quality solutions.

\begin{table}
    \footnotesize
    \centering
    \caption{Summary of experiments on T instances - Part \RNum{1}}
    \label{tab:T_results_summary1}
    
    \begin{tabular}{lrrrrrrrrr}
    \toprule
    \multicolumn{1}{c}{\multirow{3}{*}{\# Nodes}} & \multicolumn{1}{c}{\multirow{1}{*}{NDA}} & \multicolumn{2}{c}{$\text{K-Means-10}$}              & \multicolumn{2}{c}{VROOM}& \multicolumn{4}{c}{Best of 10 runs}      \\
    \cmidrule(lr){3-4} \cmidrule(lr){5-6} \cmidrule(lr){7-10}
    \multicolumn{1}{c}{} & \multicolumn{1}{c}{Cost}          & \multicolumn{1}{c}{Cost} & \multicolumn{1}{c}{Time} & \multicolumn{1}{c}{Cost} & \multicolumn{1}{c}{Time} & \multicolumn{1}{c}{$\text{GANCP}^+$} & \multicolumn{1}{c}{$g_N$} & \multicolumn{1}{c}{$g_K$} & \multicolumn{1}{c}{$g_V$} \\
     & & & \multicolumn{1}{c}{(sec)} & & \multicolumn{1}{c}{(sec)} & & \multicolumn{1}{c}{(\%)} & \multicolumn{1}{c}{(\%)} & \multicolumn{1}{c}{(\%)}\\
    \midrule
    {[}100,200{]}        & 29675.75           & 31643.21 & 3.42     & 29498.94 & 1.66     & 29321.03   & -1.20     & -7.34     & -0.60     \\
    {[}201,300{]}        & 41068.73           & 48338.13 & 8.45     & 41080.90 & 4.58     & 40651.23   & -1.02     & -15.90    & -1.05     \\
    {[}301,400{]}        & 48829.56           & 57748.89 & 11.99    & 48987.85 & 9.85     & 48434.45   & -0.81     & -16.13    & -1.13     \\
    {[}401,500{]}        & 58144.47           & 78762.95 & 14.01    & 58326.76 & 22.28    & 57719.28   & -0.73     & -26.72    & -1.04     \\
    {[}501,600{]}        & 75715.25           & 102142.93& 26.12    & 75692.72 & 48.52    & 75129.11   & -0.77     & -26.45    & -0.74     \\
    {[}601,700{]}        & 78428.69           & 119417.80& 32.54    & 78687.05 & 102.69   & 77930.80   & -0.63     & -34.74    & -0.96     \\
    {[}701,800{]}        & 101372.52          & 138626.28& 51.48    & 101334.56& 155.25   & 100516.96  & -0.84     & -27.49    & -0.81     \\
    {[}801,900{]}        & 101151.53          & 140426.73& 50.61    & 101510.61& 228.24   & 100544.05  & -0.60     & -28.40    & -0.95     \\
    {[}901,1000{]}       & 117989.13          & 166262.61& 59.84    & 118042.43& 330.17   & 117118.12  & -0.74     & -29.56    & -0.78     \\
    {[}1001,1100{]}      & 125657.77          & 181945.18& 77.62    & 125837.19& 453.42   & 124800.93  & -0.68     & -31.41    & -0.82     \\
    {[}1101,1200{]}      & 123575.23          & 182073.25& 75.59    & 124239.53& 551.04   & 122851.90  & -0.59     & -32.53    & -1.12     \\
    {[}1201,1300{]}      & 141202.37          & 212741.86& 81.78    & 141516.74& 752.71   & 140339.98  & -0.61     & -34.03    & -0.83     \\
    {[}1301,1400{]}      & 155926.93          & 238823.50& 82.95    & 155932.46& 923.81   & 155101.03  & -0.53     & -35.06    & -0.53     \\
    {[}1401,1500{]}      & 167590.06          & 228078.03& 96.12    & 167286.25& 1110.83  & 166641.89  & -0.57     & -26.94    & -0.39    \\
    \bottomrule               
\end{tabular}
\end{table}

\begin{table}
    \footnotesize
    \centering
    \caption{Summary of experiments on T instances - Part \RNum{2}}
    \label{tab:T_results_summary2}
    
    \begin{tabular}{lrrrrrrrrr}
    \toprule
    \multicolumn{1}{c}{\multirow{3}{*}{\# Nodes}} & \multicolumn{2}{c}{VROOM-$l$}       & \multicolumn{7}{c}{Average of 10 runs} \\
    \cmidrule(lr){2-3} \cmidrule(lr){4-10}
    \multicolumn{1}{c}{} & \multicolumn{1}{c}{Cost} & \multicolumn{1}{c}{Time} & \multicolumn{1}{c}{GANCP} & \multicolumn{1}{c}{Time} & \multicolumn{1}{c}{$\text{GANCP}^+$} & \multicolumn{1}{c}{Time} & \multicolumn{1}{c}{$g_N$} & \multicolumn{1}{c}{$g_K$} & \multicolumn{1}{c}{$g_V$} \\
    & & \multicolumn{1}{c}{(sec)} & & \multicolumn{1}{c}{(sec)} & & \multicolumn{1}{c}{(sec)} & \multicolumn{1}{c}{(\%)} & \multicolumn{1}{c}{(\%)} & \multicolumn{1}{c}{(\%)}\\
    \midrule
    {[}100,200{]}        & 29498.94 & 1.63  & 29117.09  & 1.65  & 29506.57& 2.90  & -0.57 & -6.75 & 0.03  \\
    {[}201,300{]}        & 41080.90 & 4.24  & 40341.34  & 3.10  & 40806.03& 7.15  & -0.64 & -15.58& -0.67 \\
    {[}301,400{]}        & 48987.85 & 9.14  & 48376.34  & 5.51  & 48604.29& 11.24 & -0.46 & -15.84& -0.78 \\
    {[}401,500{]}        & 58347.64 & 15.88 & 57233.29  & 8.75  & 57929.15& 15.42 & -0.37 & -26.45& -0.68 \\
    {[}501,600{]}        & 74519.29 & 25.28 & 74756.34  & 10.61 & 75345.94& 23.19 & -0.49 & -26.23& -0.46 \\
    {[}601,700{]}        & 78745.10 & 34.38 & 78006.76  & 14.64 & 78170.21& 30.29 & -0.33 & -34.54& -0.66 \\
    {[}701,800{]}        & 101364.25& 47.82 & 99638.54  & 16.96 & 100795.98  & 41.97 & -0.57 & -27.29& -0.53 \\
    {[}801,900{]}        & 101510.61& 52.40 & 100531.98 & 21.01 & 100784.26  & 45.56 & -0.36 & -28.23& -0.72 \\
    {[}901,1000{]}       & 118045.20& 63.88 & 116908.49 & 24.98 & 117343.41  & 54.00 & -0.55 & -29.42& -0.59 \\
    {[}1001,1100{]}      & 125837.19& 80.52 & 124396.17 & 29.29 & 125074.06  & 67.32 & -0.46 & -31.26& -0.61 \\
    {[}1101,1200{]}      & 124255.41& 87.06 & 122553.92 & 36.35 & 123121.97  & 73.37 & -0.37 & -32.38& -0.90 \\
    {[}1201,1300{]}      & 141564.49& 101.18& 140751.63 & 42.02 & 140683.88  & 82.05 & -0.37 & -33.87& -0.59 \\
    {[}1301,1400{]}      & 155932.46& 108.11& 156527.15 & 47.33 & 155526.77  & 87.82 & -0.26 & -34.88& -0.26 \\
    {[}1401,1500{]}      & 167286.57& 119.85& 166412.10 & 50.08 & 166956.54  & 97.05 & -0.38 & -26.80& -0.20           \\
    \bottomrule        
    \end{tabular}
\end{table}

In many real-life applications, the geographical location of customers and demand distributions follow a historical pattern. 
Thus, our NN is advantageous in training routing problem predictions and solving hierarchical optimization problems in real time. 
Although the computational speed of $\text{GANCP}^+$ is significantly better than that of the benchmark solver, note that the executed speed can be further enhanced while maintaining the algorithm performance. 
Table \ref{tab:T_results_summary2} shows that a significant portion of the execution time is attributed to evaluating the top candidate subproblems using a CVRP solver.
This component of time can be significantly reduced by parallelizing the HGS-CVRP Solver across multiple CPU threads.
Therefore, with additional computational resources for parallelization, one may treat the total time of the algorithm as equivalent to the execution time of GANCP. 
Parallelization of the final stage also allows for executing the CVRP solver with a higher time limit, which could potentially result in improved routing decisions.
While NDA serves as an initial solution for our heuristic, there are a few instances where the algorithm yields a solution with a positive $g_N$. 
Such cases arise due to the prediction errors. 
If desired, $\text{GANCP}^+$ can be enhanced by directly incorporating the NDA solution, if feasible, as a top solution of GANCP.
In the subsequent sections, we demonstrate the robustness of our method against data variations, in comparison to VROOM. 
Additionally, we showcase the transferability of our trained NN model to the CLRP in the appendix.

\subsection{Out-of-range Data}\label{sec:out_of_range}
We further evaluate the predictive accuracy of our method for MDVRP instances with approximate subproblem sizes that fall outside the range of the trained data. 
A summary of the results from 50 randomly generated Out-of-range (O) instances with $\frac{N}{D}\notin [50,500]$ is presented in Table \ref{tab:O_results_summary}. 
The results indicate that our method outperforms VROOM for problems with $N\in[101,250]$. 
This is likely due to the model's exposure to some out-of-range VRP instances during the stepwise training procedure and its ability to learn from them. 
Our method fails to outperform VROOM for relatively smaller and larger problems in the O instances dataset.
Nevertheless, the results can be considered satisfactory due to the relatively small positive gap.

\begin{table}
    \footnotesize
    \centering
    \caption{Summary of experiments on O instances}
    \label{tab:O_results_summary}
    \begin{tabular}{lrrrrrrrr}
    \toprule
    \multicolumn{1}{c}{\multirow{3}{*}{\#   Nodes}} & \multicolumn{2}{c}{VROOM}            & \multicolumn{4}{c}{Average of 10 runs}        & \multicolumn{2}{c}{Best of 10 runs} \\
    \cmidrule(lr){2-3} \cmidrule(lr){4-7} \cmidrule(lr){8-9}
    \multicolumn{1}{c}{}            & \multicolumn{1}{c}{Cost} & \multicolumn{1}{c}{Time}  & \multicolumn{1}{c}{GANCP} & \multicolumn{1}{c}{$\text{GANCP}^+$} & \multicolumn{1}{c}{Time}  & \multicolumn{1}{c}{$g_V$} & \multicolumn{1}{c}{$\text{GANCP}^+$} & \multicolumn{1}{c}{$g_V$} \\
    \multicolumn{1}{c}{}            & \multicolumn{1}{c}{}     & \multicolumn{1}{c}{(sec)} & \multicolumn{1}{c}{}        & \multicolumn{1}{c}{}     & \multicolumn{1}{c}{(sec)} & \multicolumn{1}{c}{(\%)} & \multicolumn{1}{c}{}     & \multicolumn{1}{c}{(\%)} \\
    \midrule
    {[}50,100{]}    & 16371.24 & 0.23      & 16328.06    & 16513.03 & 2.15      & 0.94     & 16467.98 & 0.66     \\
    {[}101,150{]}   & 22040.55 & 0.80      & 22021.44    & 22027.67 & 3.65      & -0.05    & 21930.31 & -0.49    \\
    {[}151,200{]}   & 25161.54 & 1.74      & 24994.17    & 25268.67 & 4.92      & 0.48     & 25108.81 & -0.18    \\
    {[}201,250{]}   & 35356.99 & 4.21      & 35150.75    & 35246.59 & 6.55      & -0.33    & 34999.57 & -1.03    \\
    {[}1000,1500{]} & 222183.17& 330.73    & 220043.57   & 223942.65& 56.88     & 0.77     & 223747.96& 0.68               \\
    \bottomrule      
    \end{tabular}
\end{table}

\subsection{Out-of-distribution data}\label{sec:out_of_distribution}
The robustness of our method in association with the trained NN model is tested for multiple combinations of coordinates and demand distributions. \cite{uchoa2017new} uses similar customer positioning and demand distributions to generate a class of VRP benchmark instances.
The positioning of customers is determined using three different rules, namely Random (R), Clustered (C), and Random-Clustered (RC), as described in \cite{solomon1987algorithms} VRPTW instances. 
In C, a few randomly positioned customers $s \in \text{UD}[3,8]$ act as cluster seeds to attract $N - s$ customers with an exponential decay. 
In RC, half of the customers are randomly positioned, and the remaining are clustered as in C. 
Two different types of demand distributions such as Unitary (U), where each demand is equal to 1, and Quadrant-Dependent (Q), where demand belongs to $\text{UD}[1,50]$ for even quadrant and $\text{UD}[51,100]$ for odd quadrant customers, respectively, are considered for evaluation. 
Our T instances were generated using R customer positioning within the grid and demand from $\text{UD}[1,100]$. 
Depots are assumed to take a random position on the grid in all cases. 
Using the above rules, eight classes of out-of-distribution MDVRP instances are generated, as shown in Table \ref{tab:DistributionsOfD}.

\begin{table}
    \footnotesize
    \centering
    \caption{D instances generated using multiple combinations of rules}
    \label{tab:DistributionsOfD}
    \begin{tabular}{ c c c }
    \toprule
    Instance Class & Customer Positioning & Demand Distribution \\ 
    \midrule
    D1     & C    & UD{[}1,100{]}\\
    D2     & RC   & UD{[}1,100{]}\\
    D3     & R    & U   \\
    D4     & R    & Q   \\
    D5     & C    & U   \\
    D6     & C    & Q   \\
    D7     & RC   & U   \\
    D8     & RC   & Q             \\
    \bottomrule        
    \end{tabular}
\end{table}

\begin{table}
    \footnotesize
    \centering
    \caption{Summary of experiments on D instances}
    \label{tab:D_results_summary}
    \begin{tabular}{crrrrrrrr}
    \toprule
    \multicolumn{1}{c}{\multirow{1}{*}{Instance}} & \multicolumn{2}{c}{VROOM}            & \multicolumn{4}{c}{Average of 10 runs}        & \multicolumn{2}{c}{Best of 10 runs} \\
    \cmidrule(lr){2-3} \cmidrule(lr){4-7} \cmidrule(lr){8-9}
    \multicolumn{1}{c}{\multirow{1}{*}{Class}}            & \multicolumn{1}{c}{Cost} & \multicolumn{1}{c}{Time}  & \multicolumn{1}{c}{GANCP} & \multicolumn{1}{c}{$\text{GANCP}^+$} & \multicolumn{1}{c}{Time}  & \multicolumn{1}{c}{$g_V$} & \multicolumn{1}{c}{$\text{GANCP}^+$} & \multicolumn{1}{c}{$g_V$} \\
    \multicolumn{1}{c}{}            & \multicolumn{1}{c}{}     & \multicolumn{1}{c}{(sec)} & \multicolumn{1}{c}{}        & \multicolumn{1}{c}{}     & \multicolumn{1}{c}{(sec)} & \multicolumn{1}{c}{(\%)} & \multicolumn{1}{c}{}     & \multicolumn{1}{c}{(\%)} \\
    \midrule
    D1              & 77202.13 & 389.16    & 79405.89    & 77606.12 & 39.69     & 0.62     & 77347.84 & 0.20     \\
    D2              & 80189.65 & 173.94    & 79777.80    & 80365.79 & 29.34     & 0.08     & 80178.28 & -0.19    \\
    D3              & 82998.27 & 235.70    & 86458.29    & 83043.41 & 35.46     & 0.06     & 82900.18 & -0.15    \\
    D4              & 85577.51 & 288.23    & 85711.91    & 85326.44 & 40.43     & -0.26    & 85002.43 & -0.69    \\
    D5              & 54606.09 & 106.43    & 59134.12    & 55126.99 & 22.23     & 1.28     & 54847.38 & 0.56     \\
    D6              & 63872.70 & 330.80    & 67575.29    & 64184.02 & 37.20     & 0.49     & 63903.97 & 0.06     \\
    D7              & 77442.77 & 98.38     & 81024.07    & 77722.09 & 26.95     & 0.49     & 77527.93 & 0.23     \\
    D8              & 67933.00 & 126.03    & 68419.02    & 68330.90 & 22.05     & 0.80     & 68060.73 & 0.17    \\
    \bottomrule
    \end{tabular}
\end{table}

A performance summary of $\text{GANCP}^+$ for the eight instance classes, each of different distributions, can be found in Table \ref{tab:D_results_summary}. 
Given that our NN model is trained over uniform distributions, the results for out-of-distribution instances are satisfactory, despite the positive performance gap compared to the solver in some cases. 
A noticeable difference is observed for D4 instances, where customer positioning and demand distribution resemble our T instances with minor modifications. 
Here, the average result of our heuristics in 10 runs outperforms VROOM. 
Similarly, the maximum deviation from optimality is visible for instances in class D5 where customer locations are purely clustered, and all demands are assumed to be unitary. 
Moreover, we still retain a considerable computational advantage for out-of-distribution data, while maintaining the ability to achieve satisfactory results.

\begin{table}
    \footnotesize
    \centering
    \caption{Summary of experiments on D instances after retraining the NN with partial data}
    \label{tab:D_tuned_results_summary}
    
    \begin{tabular}{crrrrrr}
    \toprule
    \multicolumn{1}{c}{\multirow{1}{*}{Instance}} & \multicolumn{1}{c}{\multirow{1}{*}{NN Train}} & \multicolumn{3}{c}{Average of 10 runs}      & \multicolumn{2}{c}{Best of 10 runs}\\
    \cmidrule(lr){3-5} \cmidrule(lr){6-7}
    \multicolumn{1}{c}{\multirow{1}{*}{Class}}      & \multicolumn{1}{c}{MAPE}     & \multicolumn{1}{c}{GANCP} & \multicolumn{1}{c}{$\text{GANCP}^+$} & \multicolumn{1}{c}{$g_V$} & \multicolumn{1}{c}{$\text{GANCP}^+$} & \multicolumn{1}{c}{$g_V$} \\
    \multicolumn{1}{c}{}       & \multicolumn{1}{c}{(\%)}     & \multicolumn{1}{c}{}      & \multicolumn{1}{c}{}       & \multicolumn{1}{c}{(\%)} & \multicolumn{1}{c}{}       & \multicolumn{1}{c}{(\%)} \\
    \midrule
    D1      & 2.96      & 80415.02  & 77698.81& 0.73  & 77321.69 & 0.06  \\
    D2      & 2.63      & 80811.31  & 80389.15 & 0.12  & 80122.80 & -0.24 \\
    D3      & 2.33      & 85363.50  & 83021.85 & 0.05  & 82890.92 & -0.17 \\
    D4      & 2.47      & 85458.10  & 85388.37 & -0.20 & 85011.12 & -0.73 \\
    D5      & 2.77      & 58020.85  & 55080.95 & 1.16  & 54793.58 & 0.46  \\
    D6      & 3.25      & 67482.74  & 64234.44 & 0.88  & 63825.88 & -0.06 \\
    D7      & 2.39      & 79556.12  & 77695.78 & 0.44  & 77524.72 & 0.15  \\
    D8      & 2.65      & 67549.30  & 68311.91 & 0.71  & 68030.88 & 0.10 \\
    \bottomrule
    \end{tabular}
\end{table}

In Table \ref{tab:D_tuned_results_summary}, we present the results of $\text{GANCP}^+$ after retraining the NN parameters with partial data from untrained distributions. 
The training data for each distribution class comprises 32,000 CVRP instances following that distribution, in addition to Step 4 instances of Phase \RNum{3} generated from random MDVRP instances.
This constitutes only 50\% of the train/validation data compared to our main experiments. 
Before training, the NN model is initialized with the parameter weights learned from Phase \RNum{3}.
We report the training accuracy observed at convergence for each training experiment.
In cases where the train MAPE is high, such as D1 and D6, the average performance of $\text{GANCP}^+$ decreases compared to Table \ref{tab:D_results_summary}. 
However, fine-tuning the NN model with partial train data leads to improvements in multiple distribution classes.
Remarkably, in all cases, the best cost from 10 experiments improved in comparison to the experiments with pretrained NN parameters.
This underscores the potential to enhance the average performance accuracy of our solution framework by incorporating additional data, in the event of a change in data trend.

\section{Conclusions} \label{sec:conclusion}

In this work, we propose a genetic algorithm to solve the MDVRP based on a decomposition approach where the fitness of a chromosome is evaluated using a neural network-based cost predictor instead of solving the optimization subproblems.
Our method $\text{GANCP}$ in conjunction with a hybrid genetic algorithm for post-processing is referred to as $\text{GANCP}^+$.
Numerical experiments show that our approach is effective in obtaining good solutions for large-scale data with significant computational advantages in comparison to VROOM, an effective open-source generic VRP solver.
We further benchmark our results using simple decision rules that decompose the problem into CVRP subproblems and also discuss the drawbacks of employing such strategies in practice. 
The accuracy of the prediction model depends on the similarity of training and testing data distributions and problem size. 
Our experiments on out-of-distribution and out-of-range instances show that the results are satisfactory for practical execution with a significant advantage in computational time. 
A direct extension of this work is to consider route duration constraints for the MDVRP. 
Additionally, a supervised learning model can be developed to more accurately detect infeasible solutions, thereby enhancing the performance of our heuristic. 
The idea of subproblem cost prediction is a promising research direction and can be extended to multiple optimization problems such as the Orienteering Problem, Prize-Collecting VRP, Periodic VRP, VRP with time windows, Split Delivery VRP, and CLRP. 
We verify this idea by using the existing trained NN model to evaluate the CLRP benchmark datasets.
It further validates the transferability of our heuristic and pre-trained NN model for a more generalized hierarchical VRP and still leads to satisfactory results.

%% file: tables/hvrp_table_cvrp.tex
\begin{table}
    \footnotesize
    \centering
    \caption{Average Results of 20,000 CVRP Test Instances}
    \label{tab:cvrp_avg}
    \begin{tabular}{rrrrrr}
        \toprule
        \multicolumn{1}{c}{\multirow{2}{*}{\# Nodes}} & \multicolumn{5}{c}{$|g_H|$ (\%)}  \\
        \cmidrule(lr){2-6}
        \multicolumn{1}{c}{}   & \multicolumn{1}{c}{\citet{robust1990implementing}} & \multicolumn{1}{c}{\citet{figliozzi2008planning}} & \multicolumn{1}{c}{Phase I} & \multicolumn{1}{c}{Phase II} & \multicolumn{1}{c}{Phase III} \\
        \midrule
        {[}50,100{]} & 27.01 & 11.09 & 1.99& 3.58 & 3.15  \\
        {[}101,150{]}& 22.22 & 11.08 & 1.40& 2.65 & 2.61  \\
        {[}151,200{]}& 17.09 & 11.44 & 1.16& 2.25 & 2.42  \\
        {[}201,250{]}& 13.13 & 11.94 & 0.99& 2.00 & 2.16  \\
        {[}251,300{]}& 11.79 & 11.75 & 0.88& 1.90 & 2.04  \\
        {[}301,350{]}& 11.92 & 11.97 & 0.83& 1.82 & 2.01  \\
        {[}351,400{]}& 13.49 & 12.23 & 0.77& 1.85 & 1.97  \\
        {[}401,450{]}& 15.54 & 12.15 & 0.72& 1.81 & 1.96  \\
        {[}451,500{]}& 19.25 & 12.18 & 0.72& 1.88 & 1.92    \\
        \bottomrule       
    \end{tabular}
\end{table}

%% file: tables/hvrp_table_cordeau.tex
\begin{table}
    \footnotesize
    \centering
    \caption{Evaluation of Cordeau Instances}
    \label{tab:cordeau}
    \begin{tabular}{ l r rr rrr rrr rr}
    \toprule
    
     & & \multicolumn{2}{c}{NDA} & \multicolumn{3}{c}{VROOM} & \multicolumn{2}{c}{$\text{GANCP}$} & \multicolumn{3}{c}{$\text{GANCP}^+$} \\
     \cmidrule(lr){3-4}
     \cmidrule(lr){5-7}
     \cmidrule(lr){8-9}  
     \cmidrule(lr){10-12}
     Instance   & Best    & Cost & Gap & Cost    & Time & Gap   & NN Cost & Time  &   Cost  & Time   &  Gap \\
                & Known   &         & (\%) &         & (sec)  &  (\%)&         & (sec) &  & (sec) & (\%)\\
     \midrule 
     p01-n50-d4                                    & 576.87                    & 609.24                   & 5.61                         & 576.87                   & 0.15                     & 0.00                         & 697.00                   & 1.64                     & 586.39                   & 3.64                     & 1.65                         \\
     p02-n50-d4                                    & 473.53                    & 507.01                   & 7.07                         & 476.66                   & 0.08                     & 0.66                         & 593.56                   & 1.61                     & 495.38                   & 3.61                     & 4.61                         \\
     p03-n75-d5                                    & 641.19                    & 681.72                   & 6.32                         & 647.89                   & 0.28                     & 1.04                         & 844.65                   & 2.23                     & 662.14                   & 4.73                     & 3.27                         \\
     p04-n100-d2                                   & 1001.04                   & 1016.64                  & 1.56                         & 1013.50                  & 0.52                     & 1.24                         & 1121.68                  & 1.16                     & 1013.37                  & 2.16                     & 1.23                         \\
     p05-n100-d2                                   & 750.03                    & 774.33                   & 3.24                         & 759.05                   & 0.52                     & 1.20                         & 876.46                   & 1.24                     & 763.90                   & 2.24                     & 1.85                         \\
     p06-n100-d3                                   & 876.50                    & 890.35                   & 1.58                         & 884.67                   & 0.55                     & 0.93                         & 1026.63                  & 1.60                     & 885.92                   & 3.11                     & 1.07                         \\
     p07-n100-d4                                   & 881.97                    & -                        & -                            & 901.82                   & 0.48                     & 2.25                         & 1102.56                  & 1.24                     & 916.65                   & 3.24                     & 3.93                        \\     
    \bottomrule   
    \end{tabular}
\end{table}

%% file: hvrp_appendix.tex
We choose the Capacitated Location Routing Problem (CLRP) as an example to demonstrate the applicability of our decomposition heuristic with subproblem cost predictions to other HVRPs. 
CLRP is a three-level hierarchical optimization problem. 
It involves opening depots, assigning customers to the opened depots, and making vehicle routing decisions to minimize overall operational costs while fulfilling all customer demands.
One can refer to \cite{laporte1986exact} for a mathematical formulation of the CLRP.  
We consider CLRP instances where the depots and vehicles are capacitated, and opening a depot or a route (vehicle) results in depot/vehicle opening costs. 
After selecting the depots to open from a set of candidate locations, the CLRP reduces into a variant of the MDVRP. 
Here, the main difference in the MDVRP variant from our T instances lies in the capacitated depots and the ability to add multiple routes (i.e., utilize more vehicles), each incurring an opening cost.

\subsection{Solution Methodology} \label{sec:clrp}

\begin{figure}
    \centering
    \begin{tikzpicture}
        \matrix [matrix of nodes] (p) %
        {
           $d$: & 1 & 2 & 3 & 4 & 5 \\
            & |[draw]| 1 & |[draw]| 1 & |[draw]| 0 & |[draw]| 0 & |[draw]| 1 \\
        };

        \matrix [right = 1em of p, matrix of nodes] %
        {
           $i$: & 1 & 2 & 3 & 4 & 5 & 6 & 7 & 8 \\
            & |[draw]| 2 & |[draw]| 5 & |[draw]| 1 & |[draw]| 1 & |[draw]| 5 & |[draw]| 2 & |[draw]| 1 & |[draw]| 5 \\
        };
    \end{tikzpicture}
    \caption{Depot opening and customer allocation chromosomes}
    \label{fig:CLRP_chromosome}
\end{figure}
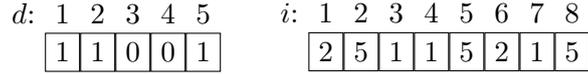

Since CLRP is a generalization of the MDVRP with additional upper level decisions, the solution methodology for MDVRP can be easily extended to this problem. 
Our GA for CLRP is extended by modifying the MDVRP chromosome representation to include both depot opening and customer assignment decisions, as shown in Figure \ref{fig:CLRP_chromosome}. 
A depot-opening chromosome is a binary vector that indicates whether a depot is open or closed, while the 
customer allocation chromosome represents the active depot assigned to a customer. 
For an allocation to be valid, the corresponding depot must be open. 
Following the GA methodology for MDVRP, each parent is selected using binary tournament selection, and crossovers are applied to the customer allocation chromosome. 
After determining all customer assignments for a child, only the depots corresponding to the allocated customers are opened on the depot chromosome. 
A similar approach can be utilized to determine the solution representation after mutations on the customer allocation chromosome. 
In each generation of the GA, a mutation of targeted customer assignments is performed to the nearest depot or the nearest neighbor's nearest depot. 
This applies to approximately $10\%$ of the candidates determined by tournament selection from the population. 
Note that since this experiment is a proof of concept for our hierarchical problem-solving approach, we do not explore advanced location searches or mutations. 

While the vehicles are capacitated in a CLRP, there is no restriction on the number of vehicles that can be utilized from a depot.
Consequently, infeasibility in a CLRP solution may stem not from the vehicle capacity constraints but rather from depot capacity constraints.
Similar to the MDVRP methodology, infeasible candidates are repaired with a probability by random assignment to available and open depots. 
During the initial population generation, we ensure that the open depots can satisfy the total demand for all customers. 
Each CLRP chromosome is identical to an MDVRP solution once the locations to be opened are determined. 
CVRP subproblems arising from the solution decompositions can be passed to the NN model in batches to obtain cost predictions. 
To assess the quality of a CLRP chromosome, subproblem routing costs and diversity scores alone are insufficient. 
Depot opening costs, an approximate estimate of vehicle opening costs (based on a lower bound on the number of vehicles), total routing cost, and solution diversity are all taken into consideration.

Our computational study focuses on the transferability of the GANCP framework and NN model from MDVRP to the CLRP. 
Therefore, we use the NN parameter weights from Phase \RNum{3} experiments in determining the solutions to CLRP benchmark datasets. 
This ensures that our approach is applicable to CLRP without requiring computational effort in training a problem-specific NN model.

\subsection{Computational Study for CLRP} \label{sec:comp_exp2}

Our modified genetic algorithm for CLRP is carried out by using the same set of trained parameters $\theta$ as in the MDVRP experiments. 
Note that some local searches, such as random mutations, are not considered in this evaluation of CLRP benchmark datasets. 
We focus on our main goal of verifying the applicability of decomposition evaluation through cost prediction and therefore do not investigate additional solution improvement strategies for the CLRP. 
We analyze the performance of our modified heuristic in two popular CLRP benchmark sets, \cite{tuzun1999two} and \cite{barreto2004analise}. 
\cite{tuzun1999two} contains $36$ instances with $N \in \{100,150,200\}$ customers, $D \in \{10,20\}$ uncapacitated depots and capacitated vehicles. 
Customers are distributed in varying densities, and the clusters are sized differently in these instances. 
\cite{barreto2004analise} instances are derived from the CVRP instances with up to $150$ customers and $10$ depots.
The route (vehicle) opening costs are ignored in these instances. 

Table \ref{tab:clrp_tuzun_detailed} and Table \ref{tab:clrp_barreto_detailed} in Appendix \ref{sec:detailed_results} demonstrate the performance of our method for the CLRP benchmark datasets. 
We report the best-known solutions for each instance by examining multiple research studies in the CLRP literature. 
Furthermore, we compare our $\text{GANCP}^+$ performance with the Tree-Based Search Algorithm (TBSA), as it is one of the most effective heuristics for CLRP. 
While TBSA has four variants, we report the results of both the fast and quality-oriented variants. 
We also compare the results of Hybrid Adaptive Large Neighborhood Search (Hybrid ALNS) which demonstrates improved performance for a few CLRP instances. 
Note that the execution time for TBSA variants and Hybrid ALNS has been directly taken from the research works of \citet{schneider2019large} and \citet{akpunar2021hybrid}, respectively. 
Our results show that the NN model provides approximate ranking estimates for the decompositions leading to good solutions, even when the prediction model has been trained on a different distribution of CVRPs.
Notably, $\text{GANCP}^+$ achieves proven optimal solutions in a few instances.

%% file: tables/hvrp_table_T.tex
\begin{longtable}{lrrrrrrrrrrrrrrrrrr}
    \caption{Results of T instances}
    \label{tab:T_results}
    \endfirsthead
    \endhead
    \toprule
    \multicolumn{1}{c}{\multirow{3}{*}{Instance}} & \multicolumn{1}{c}{\multirow{2}{*}{NDA}} & \multicolumn{2}{c}{$\text{K-Means-10}$}              & \multicolumn{2}{c}{VROOM}                 & \multicolumn{2}{c}{VROOM-$l$}            & \multicolumn{7}{c}{Average of 10 runs}  & \multicolumn{4}{c}{Best of 10 runs}        \\
    \cmidrule(lr){3-4} \cmidrule(lr){5-6} \cmidrule(lr){7-8} \cmidrule(lr){9-15} \cmidrule(lr){16-19}
    \multicolumn{1}{c}{}                & \multicolumn{1}{c}{\multirow{1}{*}{Cost}}                & \multicolumn{1}{c}{Cost} & \multicolumn{1}{c}{Time} & \multicolumn{1}{c}{Cost} & \multicolumn{1}{c}{Time} & \multicolumn{1}{c}{Cost} & \multicolumn{1}{c}{Time} & \multicolumn{1}{c}{NN Cost} & \multicolumn{1}{c}{Time} & \multicolumn{1}{c}{$\text{GANCP}^+$} & \multicolumn{1}{c}{Time} & \multicolumn{1}{c}{$g_N$} & \multicolumn{1}{c}{$g_K$} & \multicolumn{1}{c}{$g_V$} & \multicolumn{1}{c}{$\text{GANCP}^+$} & \multicolumn{1}{c}{$g_N$} & \multicolumn{1}{c}{$g_K$} & \multicolumn{1}{c}{$g_V$} \\
    & & & \multicolumn{1}{c}{(sec)} & & \multicolumn{1}{c}{(sec)} & & \multicolumn{1}{c}{(sec)} & & \multicolumn{1}{c}{(sec)}& & \multicolumn{1}{c}{(sec)} & \multicolumn{1}{c}{(\%)} & \multicolumn{1}{c}{(\%)} & \multicolumn{1}{c}{(\%)} & & \multicolumn{1}{c}{(\%)} & \multicolumn{1}{c}{(\%)} & \multicolumn{1}{c}{(\%)} \\
    \midrule 
    T001-n100-d2     & 17390.51         & 17875.93       & 9.14           & 17166.17       & 0.50           & 17166.17       & 0.49           & 16553.16          & 1.01           & 17315.14         & 2.01           & -0.43           & -3.14           & 0.87            & 17183.24         & -1.19           & -3.87           & 0.10            \\
\bottomrule
\end{longtable}

%% file: tables/hvrp_table_O.tex
\begin{longtable}{lrrrrrrrrrrrr}
    \caption{Results of O instances}
    \label{tab:O_results}
    \endfirsthead
    \endhead
    \toprule
        \multicolumn{1}{c}{\multirow{3}{*}{Instance}} & \multicolumn{2}{c}{VROOM}                            & \multicolumn{5}{c}{Average of 10 runs}                                                                                                        & \multicolumn{5}{c}{Best of 10 runs}                                                                                                           \\
        \cmidrule(lr){2-3} \cmidrule(lr){4-8} \cmidrule(lr){9-13}
        \multicolumn{1}{c}{}                          & \multicolumn{1}{c}{Cost} & \multicolumn{1}{c}{Time}  & \multicolumn{1}{c}{GANCP} & \multicolumn{1}{c}{Time}  & \multicolumn{1}{c}{$\text{GANCP}^+$} & \multicolumn{1}{c}{Time}  & \multicolumn{1}{c}{$g_V$} & \multicolumn{1}{c}{GANCP} & \multicolumn{1}{c}{Time}  & \multicolumn{1}{c}{$\text{GANCP}^+$} & \multicolumn{1}{c}{Time}  & \multicolumn{1}{c}{$g_V$} \\
        \multicolumn{1}{c}{}                          & \multicolumn{1}{c}{}     & \multicolumn{1}{c}{(sec)} & \multicolumn{1}{c}{}        & \multicolumn{1}{c}{(sec)} & \multicolumn{1}{c}{}          & \multicolumn{1}{c}{(sec)} & \multicolumn{1}{c}{(\%)} & \multicolumn{1}{c}{}        & \multicolumn{1}{c}{(sec)} & \multicolumn{1}{c}{}          & \multicolumn{1}{c}{(sec)} & \multicolumn{1}{c}{(\%)} \\
        \midrule
        O01-n55-d2                                    & 13046.00                 & 0.11                      & 13120.58                    & 0.86                      & 13119.74                      & 1.87                      & 0.57                    & 13120.58                    & 2.00                      & 13119.74                      & 3.02                      & 0.57                    \\
\bottomrule
\end{longtable}

%% file: tables/hvrp_table_D.tex
\begin{longtable}{lrrrrrrrrrrrr}
    \caption{Results of D instances}
    \label{tab:D_results}
    \endfirsthead
    \endhead
    \toprule
    \multicolumn{1}{c}{\multirow{3}{*}{Instance}} & \multicolumn{2}{c}{VROOM}                            & \multicolumn{5}{c}{Average of 10 runs}                                                                                                        & \multicolumn{5}{c}{Best of 10 runs}                                                                                                           \\
        \cmidrule(lr){2-3} \cmidrule(lr){4-8} \cmidrule(lr){9-13}
        \multicolumn{1}{c}{}                          & \multicolumn{1}{c}{Cost} & \multicolumn{1}{c}{Time}  & \multicolumn{1}{c}{GANCP} & \multicolumn{1}{c}{Time}  & \multicolumn{1}{c}{$\text{GANCP}^+$} & \multicolumn{1}{c}{Time}  & \multicolumn{1}{c}{$g_V$} & \multicolumn{1}{c}{GANCP} & \multicolumn{1}{c}{Time}  & \multicolumn{1}{c}{$\text{GANCP}^+$} & \multicolumn{1}{c}{Time}  & \multicolumn{1}{c}{$g_V$} \\
        \multicolumn{1}{c}{}                          & \multicolumn{1}{c}{}     & \multicolumn{1}{c}{(sec)} & \multicolumn{1}{c}{}        & \multicolumn{1}{c}{(sec)} & \multicolumn{1}{c}{}          & \multicolumn{1}{c}{(sec)} & \multicolumn{1}{c}{(\%)} & \multicolumn{1}{c}{}        & \multicolumn{1}{c}{(sec)} & \multicolumn{1}{c}{}          & \multicolumn{1}{c}{(sec)} & \multicolumn{1}{c}{(\%)} \\
        \midrule
        D1-n105-d2                                    & 13966.79                 & 0.37                      & 13652.76                  & 1.09                      & 14272.19                      & 2.09                      & 2.19                    & 13392.72                  & 1.14                      & 14156.04                      & 2.14                      & 1.35                    \\
\bottomrule
\end{longtable}

%% file: tables/hvrp_table_D_tuned.tex
\begin{longtable}{lcrrrrr}
    \caption{Results of D instances after retraining NN with partial data}
    \label{tab:D_tuned_results}
    \endfirsthead
    \endhead
    \toprule
        \multicolumn{1}{c}{\multirow{3}{*}{Instance}} & \multicolumn{1}{c}{NN Train} & \multicolumn{3}{c}{Average of 10 runs}                                            & \multicolumn{2}{c}{Best Cost}                
        \\
        \cmidrule(lr){3-5} \cmidrule(lr){6-7}
        \multicolumn{1}{c}{}                          & \multicolumn{1}{c}{MAPE}     & \multicolumn{1}{c}{GANCP} & \multicolumn{1}{c}{$\text{GANCP}^+$} & \multicolumn{1}{c}{$g_V$} & \multicolumn{1}{c}{$\text{GANCP}^+$} & \multicolumn{1}{c}{$g_V$} 
        \\
        \multicolumn{1}{c}{}                          & \multicolumn{1}{c}{(\%)}     & \multicolumn{1}{c}{}      & \multicolumn{1}{c}{}       & \multicolumn{1}{c}{(\%)} & \multicolumn{1}{c}{}       & \multicolumn{1}{c}{(\%)} \\
        \midrule
        D1-n105-d2                                    & \multirow{10}{*}{2.96}       & 13942.63                  & 14241.61                   & 1.97                     & 14003.42                   & 0.26                     \\
        D1-n237-d3                                    &                              & 30649.97                  & 30842.83                   & 1.33                     & 30521.95                   & 0.28                     \\
        D1-n255-d2                                    &                              & 65665.88                  & 63684.04                   & -0.39                    & 63669.52                   & -0.41                    \\
        D1-n350-d4                                    &                              & 35521.05                  & 35840.60                   & -0.24                    & 35703.83                   & -0.62                    \\
        D1-n416-d8                                    &                              & 49087.18                  & 45652.11                   & 1.96                     & 45063.58                   & 0.64                     \\
\bottomrule
\end{longtable}

%% file: tables/hvrp_table_CLRP_detailed.tex
\begin{longtable}{r r rr rr r rr rrrr rr r}
    \caption{Detailed results of Tuzun-Burke CLRP Instances}
    \label{tab:clrp_tuzun_detailed}
    \endfirsthead
    \endhead
    \toprule
    \multicolumn{1}{c}{\multirow{4}{*}{Instance}} & \multicolumn{1}{c}{\multirow{3}{*}{Best}} & \multicolumn{2}{c}{\multirow{2}{*}{$\text{TBSA}_{\text{speed}}$}}   & \multicolumn{2}{c}{\multirow{2}{*}{$\text{TBSA}_{\text{qual}}$}}    & \multicolumn{1}{c}{\multirow{3}{*}{$\bar{\text{TBSA}}$}} & \multicolumn{2}{c}{\multirow{2}{*}{Hybrid ALNS}}    & \multicolumn{5}{c}{$\text{GANCP}^+$}  \\
    \cmidrule(lr){10-14}

    \multicolumn{1}{c}{}                          & \multicolumn{1}{c}{\multirow{3}{*}{Known}}                     & \multicolumn{2}{c}{}                                & \multicolumn{2}{c}{}                                & \multicolumn{1}{c}{}                                             & \multicolumn{2}{c}{}                                & \multicolumn{1}{c}{\multirow{2}{*}{Best Cost}} & \multicolumn{1}{c}{\multirow{2}{*}{Time}} & \multicolumn{2}{c}{Opened}            & \multicolumn{1}{c}{\multirow{2}{*}{Gap}}                          \\
    \cmidrule(lr){3-4} \cmidrule(lr){5-6} \cmidrule(lr){8-9} \cmidrule(lr){12-13}

    \multicolumn{1}{c}{}                          & \multicolumn{1}{c}{}                     & \multicolumn{1}{c}{Cost} & \multicolumn{1}{c}{Time} & \multicolumn{1}{c}{Cost} & \multicolumn{1}{c}{Time} & \multicolumn{1}{c}{}                                             & \multicolumn{1}{c}{Cost} & \multicolumn{1}{c}{Time} & \multicolumn{1}{c}{}                           & \multicolumn{1}{c}{}                      & Depots & \multicolumn{1}{c}{Vehicles} & \multicolumn{1}{c}{}                          \\
    & & & \multicolumn{1}{c}{(sec)} & & \multicolumn{1}{c}{(sec)} & & & \multicolumn{1}{c}{(sec)} & & \multicolumn{1}{c}{(sec)} & & & \multicolumn{1}{c}{(\%)}\\
\bottomrule                
\end{longtable}

\small
\textbf{Remarks for Table \ref{tab:clrp_tuzun_detailed}:}
\begin{enumerate}
    \item TBSA and Hybrid ALNS results (inclusive of run times) have been directly taken from \cite{schneider2019large} and \cite{akpunar2021hybrid}, respectively. 
    \item $\text{TBSA}_{\text{speed}}$ is the fast-variant and $\text{TBSA}_{\text{qual}}$ is the quality-oriented variant of TBSA. 
    \item $\bar{\text{TBSA}}$ shows the best result from all four variants of TBSA proposed in \cite{schneider2019large}. 
\end{enumerate}

\newpage 

\begin{longtable}{lrrrrrrrrrr}
    \caption{Detailed results of Barreto CLRP Instances}
    \label{tab:clrp_barreto_detailed}
    \endfirsthead
    \endhead
    \toprule
    \multicolumn{1}{c}{\multirow{4}{*}{Instance}} & \multicolumn{1}{c}{\multirow{3}{*}{Best}} & \multicolumn{2}{c}{\multirow{2}{*}{$\text{TBSA}_{\text{speed}}$}}   & \multicolumn{2}{c}{\multirow{2}{*}{Hybrid ALNS}}    & \multicolumn{5}{c}{$\text{GANCP}^+$}   \\
    \cmidrule(lr){7-11}
    \multicolumn{1}{c}{}                          & \multicolumn{1}{c}{\multirow{3}{*}{Known}}     & \multicolumn{2}{c}{}                                & \multicolumn{2}{c}{}                                & \multicolumn{1}{c}{\multirow{2}{*}{Best Cost}} & \multicolumn{1}{c}{\multirow{2}{*}{Time}} & \multicolumn{2}{c}{Opened}                                 & \multicolumn{1}{c}{\multirow{2}{*}{Gap }}  \\
    \cmidrule(lr){3-4} \cmidrule(lr){5-6} \cmidrule(lr){9-10}
    \multicolumn{1}{c}{}                          & \multicolumn{1}{c}{}                     & \multicolumn{1}{c}{Cost} & \multicolumn{1}{c}{Time} & \multicolumn{1}{c}{Cost} & \multicolumn{1}{c}{Time} & \multicolumn{1}{c}{}                           & \multicolumn{1}{c}{}                      & \multicolumn{1}{l}{Depots} & \multicolumn{1}{c}{Vehicles}  & \multicolumn{1}{c}{}                          \\
    & & & \multicolumn{1}{c}{(sec)} & & \multicolumn{1}{c}{(sec)} & & \multicolumn{1}{c}{(sec)} & & & \multicolumn{1}{c}{(\%)}\\
    \midrule
    Christofides69-50x5                           & 565.6                                    & 565.6                    & 1.94                     & 565.6                    & 60                       & 592.9                                          & 2.90                                      & 1                          & 5                             & 4.83                                          \\
\bottomrule      
\end{longtable}

\small
\textbf{Remarks for Table \ref{tab:clrp_barreto_detailed}:}
\begin{enumerate}
    \item TBSA and Hybrid ALNS results (inclusive of run times) have been directly taken from \cite{schneider2019large} and \cite{akpunar2021hybrid}, respectively. 
    \item $\text{TBSA}_{\text{speed}}$ is the fast-variant of the TBSA. The quality-oriented variant of TBSA, $\text{TBSA}_{\text{qual}}$,  produced the same results as $\text{TBSA}_{\text{speed}}$ with an average run time of 171.43 seconds for Barreto instances.
\end{enumerate}
\scriptsize

%% file: hvrp_main.bbl
\begin{thebibliography}{61}
\expandafter\ifx\csname natexlab\endcsname\relax\def\natexlab#1{#1}\fi
\expandafter\ifx\csname url\endcsname\relax
  \def\url#1{{\tt #1}}\fi
\expandafter\ifx\csname urlprefix\endcsname\relax\def\urlprefix{URL }\fi
\expandafter\ifx\csname urlstyle\endcsname\relax
  \expandafter\ifx\csname doi\endcsname\relax
  \def\doi#1{doi:\discretionary{}{}{}#1}\fi \else
  \expandafter\ifx\csname doi\endcsname\relax
  \def\doi{doi:\discretionary{}{}{}\begingroup \urlstyle{rm}\Url}\fi \fi

\bibitem[{Akkerman and Mes(2022)}]{akkerman2022distance}
Akkerman, F., M. Mes. 2022.
\newblock Distance approximation to support customer selection in vehicle
  routing problems.
\newblock {\it Annals of Operations Research\/}  1--29.

\bibitem[{Akpunar and Akpinar(2021)}]{akpunar2021hybrid}
Akpunar, {\"O}.~{\c{S}}., {\c{S}}. Akpinar. 2021.
\newblock A hybrid adaptive large neighbourhood search algorithm for the
  capacitated location routing problem.
\newblock {\it Expert Systems with Applications\/} {\bf 168} 114304.

\bibitem[{Arnold et~al.(2019)Arnold, Gendreau, and
  S{\"o}rensen}]{arnold2019efficiently}
Arnold, F., M. Gendreau, K. S{\"o}rensen. 2019.
\newblock Efficiently solving very large-scale routing problems.
\newblock {\it Computers \& operations research\/} {\bf 107} 32--42.

\bibitem[{Ba et~al.(2016)Ba, Kiros, and Hinton}]{ba2016layer}
Ba, J.~L., J.~R. Kiros, G.~E. Hinton. 2016.
\newblock Layer normalization.
\newblock {\it Advances in NIPS 2016 Deep Learning Symposium\/}. (p. arXiv
  preprint arXiv:1607.06450).

\bibitem[{Baldacci and Mingozzi(2009)}]{baldacci2009unified}
Baldacci, R., A. Mingozzi. 2009.
\newblock A unified exact method for solving different classes of vehicle
  routing problems.
\newblock {\it Mathematical Programming\/} {\bf 120} 347--380.

\bibitem[{Baldacci et~al.(2011)Baldacci, Mingozzi, and
  Wolfler~Calvo}]{baldacci2011exact}
Baldacci, R., A. Mingozzi, R. Wolfler~Calvo. 2011.
\newblock An exact method for the capacitated location-routing problem.
\newblock {\it Operations Research\/} {\bf 59}(5) 1284--1296.

\bibitem[{Banerjee et~al.(2022)Banerjee, Erera, and
  Toriello}]{banerjee2022fleet}
Banerjee, D., A.~L. Erera, A. Toriello. 2022.
\newblock Fleet sizing and service region partitioning for same-day delivery
  systems.
\newblock {\it Transportation Science\/} {\bf 56}(5) 1327--1347.

\bibitem[{Barreto(2004)}]{barreto2004analise}
Barreto, S. d.~S. 2004.
\newblock An{\'a}lise e modeliza{\c{c}}{\~a}o de problemas de
  localiza{\c{c}}{\~a}o-distribui{\c{c}}{\~a}o [analysis and modelling of
  location-routing problems].
\newblock {\it Unpublished doctoral dissertation, University of Aveiro, Campus
  Universit{\'a}rio de Santiago\/}  3810--193.

\bibitem[{Belenguer et~al.(2011)Belenguer, Benavent, Prins, Prodhon, and
  Calvo}]{belenguer2011branch}
Belenguer, J.-M., E. Benavent, C. Prins, C. Prodhon, R.~W. Calvo. 2011.
\newblock A branch-and-cut method for the capacitated location-routing problem.
\newblock {\it Computers \& Operations Research\/} {\bf 38}(6) 931--941.

\bibitem[{Bello et~al.(2017)Bello, Pham, Le, Norouzi, and
  Bengio}]{bello2016neural}
Bello, I., H. Pham, Q.~V. Le, M. Norouzi, S. Bengio. 2017.
\newblock Neural combinatorial optimization with reinforcement learning.
\newblock {\it 5th International Conference on Learning Representations, {ICLR}
  2017, Toulon, France, April 24-26, 2017, Workshop Track Proceedings\/}.
  OpenReview.net.
\newblock \urlprefix\url{https://openreview.net/forum?id=Bk9mxlSFx}.

\bibitem[{Bogyrbayeva et~al.(2023)Bogyrbayeva, Yoon, Ko, Lim, Yun, and
  Kwon}]{bogyrbayeva2023deep}
Bogyrbayeva, A., T. Yoon, H. Ko, S. Lim, H. Yun, C. Kwon. 2023.
\newblock A deep reinforcement learning approach for solving the traveling
  salesman problem with drone.
\newblock {\it Transportation Research Part C: Emerging Technologies\/} {\bf
  148} 103981.

\bibitem[{{CapGemini Research Institute}(2019)}]{capgemini}
{CapGemini Research Institute}. 2019.
\newblock The last-mile delivery challenge.
\newblock Tech. rep., CapGemini Research Institute.

\bibitem[{{Capital One Shopping Research}(2019)}]{capitalone}
{Capital One Shopping Research}. 2019.
\newblock Amazon logistics statistics.
\newblock Tech. rep., Capital One Shopping Research.

\bibitem[{Cappart et~al.(2023)Cappart, Ch{\'e}telat, Khalil, Lodi, Morris, and
  Veli{\v{c}}kovi{\'c}}]{cappart2023combinatorial}
Cappart, Q., D. Ch{\'e}telat, E.~B. Khalil, A. Lodi, C. Morris, P.
  Veli{\v{c}}kovi{\'c}. 2023.
\newblock Combinatorial optimization and reasoning with graph neural networks.
\newblock {\it Journal of Machine Learning Research\/} {\bf 24}(130) 1--61.

\bibitem[{Contardo et~al.(2014)Contardo, Cordeau, and
  Gendron}]{contardo2014exact}
Contardo, C., J.-F. Cordeau, B. Gendron. 2014.
\newblock An exact algorithm based on cut-and-column generation for the
  capacitated location-routing problem.
\newblock {\it INFORMS Journal on Computing\/} {\bf 26}(1) 88--102.

\bibitem[{Contardo and Martinelli(2014)}]{contardo2014new}
Contardo, C., R. Martinelli. 2014.
\newblock A new exact algorithm for the multi-depot vehicle routing problem
  under capacity and route length constraints.
\newblock {\it Discrete Optimization\/} {\bf 12} 129--146.

\bibitem[{Cordeau et~al.(1997)Cordeau, Gendreau, and Laporte}]{cordeau1997tabu}
Cordeau, J.-F., M. Gendreau, G. Laporte. 1997.
\newblock A tabu search heuristic for periodic and multi-depot vehicle routing
  problems.
\newblock {\it Networks: An International Journal\/} {\bf 30}(2) 105--119.

\bibitem[{Coupey et~al.(2023)Coupey, Nicod, and Varnier}]{vroom}
Coupey, J., J.-M. Nicod, C. Varnier. 2023.
\newblock {VROOM v1.13, Vehicle Routing Open-source Optimization Machine}.
\newblock \url{http://vroom-project.org/} (Last Accessed: October 1, 2023).

\bibitem[{de~Oliveira et~al.(2016)de~Oliveira, Enayatifar, Sadaei,
  Guimar{\~a}es, and Potvin}]{de2016cooperative}
de~Oliveira, F.~B., R. Enayatifar, H.~J. Sadaei, F.~G. Guimar{\~a}es, J.-Y.
  Potvin. 2016.
\newblock A cooperative coevolutionary algorithm for the multi-depot vehicle
  routing problem.
\newblock {\it Expert Systems with Applications\/} {\bf 43} 117--130.

\bibitem[{Ellegood et~al.(2015)Ellegood, Campbell, and
  North}]{ellegood2015continuous}
Ellegood, W.~A., J.~F. Campbell, J. North. 2015.
\newblock Continuous approximation models for mixed load school bus routing.
\newblock {\it Transportation Research Part B: Methodological\/} {\bf 77}
  182--198.

\bibitem[{Figliozzi(2008)}]{figliozzi2008planning}
Figliozzi, M.~A. 2008.
\newblock Planning approximations to the average length of vehicle routing
  problems with varying customer demands and routing constraints.
\newblock {\it Transportation Research Record\/} {\bf 2089}(1) 1--8.

\bibitem[{Franceschetti et~al.(2017)Franceschetti, Jabali, and
  Laporte}]{franceschetti2017continuous}
Franceschetti, A., O. Jabali, G. Laporte. 2017.
\newblock Continuous approximation models in freight distribution management.
\newblock {\it Top\/} {\bf 25} 413--433.

\bibitem[{Furian et~al.(2021)Furian, O'Sullivan, Walker, and
  {\c{C}}ela}]{furian2021machine}
Furian, N., M. O'Sullivan, C. Walker, E. {\c{C}}ela. 2021.
\newblock A machine learning-based branch and price algorithm for a sampled
  vehicle routing problem.
\newblock {\it OR Spectrum\/} {\bf 43} 693--732.

\bibitem[{Garn(2021)}]{garn2021balanced}
Garn, W. 2021.
\newblock Balanced dynamic multiple travelling salesmen: Algorithms and
  continuous approximations.
\newblock {\it Computers \& Operations Research\/} {\bf 136} 105509.

\bibitem[{Ghaffarinasab et~al.(2018)Ghaffarinasab, Van~Woensel, and
  Minner}]{ghaffarinasab2018continuous}
Ghaffarinasab, N., T. Van~Woensel, S. Minner. 2018.
\newblock A continuous approximation approach to the planar hub
  location-routing problem: Modeling and solution algorithms.
\newblock {\it Computers \& Operations Research\/} {\bf 100} 140--154.

\bibitem[{Golden et~al.(1977)Golden, Magnanti, and
  Nguyen}]{golden1977implementing}
Golden, B.~L., T.~L. Magnanti, H.~Q. Nguyen. 1977.
\newblock Implementing vehicle routing algorithms.
\newblock {\it Networks\/} {\bf 7}(2) 113--148.

\bibitem[{He et~al.(2016)He, Zhang, Ren, and Sun}]{he2016deep}
He, K., X. Zhang, S. Ren, J. Sun. 2016.
\newblock Deep residual learning for image recognition.
\newblock {\it Proceedings of the IEEE Conference on Computer Vision and
  Pattern Recognition\/}. 770--778.

\bibitem[{Kim et~al.(2024)Kim, Park, and Kwon}]{kim2024neural}
Kim, H., J. Park, C. Kwon. 2024.
\newblock A neural separation algorithm for the rounded capacity inequalities.
\newblock {\it INFORMS Journal on Computing\/} .

\bibitem[{Kim et~al.(2023)Kim, Park, and Park}]{kim2023learning}
Kim, M., J. Park, J. Park. 2023.
\newblock Learning to {CROSS} exchange to solve min-max vehicle routing
  problems.
\newblock {\it The Eleventh International Conference on Learning
  Representations\/}.
\newblock \urlprefix\url{https://openreview.net/forum?id=ZcnzsHC10Y}.

\bibitem[{Kool et~al.(2019)Kool, van Hoof, and Welling}]{kool2018attention}
Kool, W., H. van Hoof, M. Welling. 2019.
\newblock Attention, learn to solve routing problems!
\newblock {\it International Conference on Learning Representations\/}.
\newblock \urlprefix\url{https://openreview.net/forum?id=ByxBFsRqYm}.

\bibitem[{Kou et~al.(2023)Kou, Golden, and Bertazzi}]{kou2023improved}
Kou, S., B. Golden, L. Bertazzi. 2023.
\newblock An improved model for estimating optimal vrp solution values.
\newblock {\it Optimization Letters\/}  1--7.

\bibitem[{Kwon et~al.(2020)Kwon, Choo, Kim, Yoon, Gwon, and Min}]{kwon2020pomo}
Kwon, Y.-D., J. Choo, B. Kim, I. Yoon, Y. Gwon, S. Min. 2020.
\newblock Pomo: Policy optimization with multiple optima for reinforcement
  learning.
\newblock {\it Advances in Neural Information Processing Systems\/} {\bf 33}
  21188--21198.

\bibitem[{Laporte et~al.(1986)Laporte, Nobert, and Arpin}]{laporte1986exact}
Laporte, G., Y. Nobert, D. Arpin. 1986.
\newblock An exact algorithm for solving a capacitated location-routing
  problem.
\newblock {\it Annals of Operations Research\/} {\bf 6} 291--310.

\bibitem[{Laporte et~al.(1988)Laporte, Nobert, and
  Taillefer}]{laporte1988solving}
Laporte, G., Y. Nobert, S. Taillefer. 1988.
\newblock Solving a family of multi-depot vehicle routing and location-routing
  problems.
\newblock {\it Transportation Science\/} {\bf 22}(3) 161--172.

\bibitem[{Li et~al.(2021)Li, Yan, and Wu}]{li2021learning}
Li, S., Z. Yan, C. Wu. 2021.
\newblock Learning to delegate for large-scale vehicle routing.
\newblock {\it Advances in Neural Information Processing Systems\/} {\bf 34}
  26198--26211.

\bibitem[{Lu et~al.(2020)Lu, Zhang, and Yang}]{Lu2020A}
Lu, H., X. Zhang, S. Yang. 2020.
\newblock A learning-based iterative method for solving vehicle routing
  problems.
\newblock {\it International Conference on Learning Representations\/}.
\newblock \urlprefix\url{https://openreview.net/forum?id=BJe1334YDH}.

\bibitem[{Morabit et~al.(2021)Morabit, Desaulniers, and
  Lodi}]{morabit2021machine}
Morabit, M., G. Desaulniers, A. Lodi. 2021.
\newblock Machine-learning--based column selection for column generation.
\newblock {\it Transportation Science\/} {\bf 55}(4) 815--831.

\bibitem[{Nazari et~al.(2018)Nazari, Oroojlooy, Snyder, and
  Tak{\'a}c}]{nazari2018reinforcement}
Nazari, M., A. Oroojlooy, L. Snyder, M. Tak{\'a}c. 2018.
\newblock Reinforcement learning for solving the vehicle routing problem.
\newblock {\it Advances in Neural Information Processing Systems\/} {\bf 31}.

\bibitem[{Nicola et~al.(2019)Nicola, Vetschera, and Dragomir}]{nicola2019total}
Nicola, D., R. Vetschera, A. Dragomir. 2019.
\newblock Total distance approximations for routing solutions.
\newblock {\it Computers \& Operations Research\/} {\bf 102} 67--74.

\bibitem[{Park et~al.(2023)Park, Kwon, and Park}]{park2023learn}
Park, J., C. Kwon, J. Park. 2023.
\newblock Learn to solve the min-max multiple traveling salesmen problem with
  reinforcement learning.
\newblock {\it Proceedings of the 2023 International Conference on Autonomous
  Agents and Multiagent Systems (AAMAS ‘23)\/}. 878--886.

\bibitem[{Qi et~al.(2012)Qi, Lin, Li, and Miao}]{qi2012spatiotemporal}
Qi, M., W.-H. Lin, N. Li, L. Miao. 2012.
\newblock A spatiotemporal partitioning approach for large-scale vehicle
  routing problems with time windows.
\newblock {\it Transportation Research Part E: Logistics and Transportation
  Review\/} {\bf 48}(1) 248--257.

\bibitem[{Ramos et~al.(2020)Ramos, Gomes, and P{\'o}voa}]{ramos2020multi}
Ramos, T. R.~P., M.~I. Gomes, A.~P.~B. P{\'o}voa. 2020.
\newblock Multi-depot vehicle routing problem: a comparative study of
  alternative formulations.
\newblock {\it International Journal of Logistics Research and Applications\/}
  {\bf 23}(2) 103--120.

\bibitem[{Robust et~al.(1990)Robust, Daganzo, and
  Souleyrette~II}]{robust1990implementing}
Robust, F., C.~F. Daganzo, R.~R. Souleyrette~II. 1990.
\newblock Implementing vehicle routing models.
\newblock {\it Transportation Research Part B: Methodological\/} {\bf 24}(4)
  263--286.

\bibitem[{Saberi and Verbas(2012)}]{saberi2012continuous}
Saberi, M., {\.I}.~{\"O}. Verbas. 2012.
\newblock Continuous approximation model for the vehicle routing problem for
  emissions minimization at the strategic level.
\newblock {\it Journal of Transportation Engineering\/} {\bf 138}(11)
  1368--1376.

\bibitem[{Sadati et~al.(2021)Sadati, {\c{C}}atay, and
  Aksen}]{sadati2021efficient}
Sadati, M. E.~H., B. {\c{C}}atay, D. Aksen. 2021.
\newblock An efficient variable neighborhood search with tabu shaking for a
  class of multi-depot vehicle routing problems.
\newblock {\it Computers \& Operations Research\/} {\bf 133} 105269.

\bibitem[{Schneider and L{\"o}ffler(2019)}]{schneider2019large}
Schneider, M., M. L{\"o}ffler. 2019.
\newblock Large composite neighborhoods for the capacitated location-routing
  problem.
\newblock {\it Transportation Science\/} {\bf 53}(1) 301--318.

\bibitem[{Solomon(1987)}]{solomon1987algorithms}
Solomon, M.~M. 1987.
\newblock Algorithms for the vehicle routing and scheduling problems with time
  window constraints.
\newblock {\it Operations Research\/} {\bf 35}(2) 254--265.

\bibitem[{Stroh et~al.(2022)Stroh, Erera, and Toriello}]{stroh2022tactical}
Stroh, A.~M., A.~L. Erera, A. Toriello. 2022.
\newblock Tactical design of same-day delivery systems.
\newblock {\it Management Science\/} {\bf 68}(5) 3444--3463.

\bibitem[{Tuzun and Burke(1999)}]{tuzun1999two}
Tuzun, D., L.~I. Burke. 1999.
\newblock A two-phase tabu search approach to the location routing problem.
\newblock {\it European Journal of Operational Research\/} {\bf 116}(1) 87--99.

\bibitem[{Uchoa et~al.(2017)Uchoa, Pecin, Pessoa, Poggi, Vidal, and
  Subramanian}]{uchoa2017new}
Uchoa, E., D. Pecin, A. Pessoa, M. Poggi, T. Vidal, A. Subramanian. 2017.
\newblock New benchmark instances for the capacitated vehicle routing problem.
\newblock {\it European Journal of Operational Research\/} {\bf 257}(3)
  845--858.

\bibitem[{Varol et~al.(2024)Varol, {\"O}zener, and Albey}]{varol2024neural}
Varol, T., O.~{\"O}. {\"O}zener, E. Albey. 2024.
\newblock Neural network estimators for optimal tour lengths of traveling
  salesperson problem instances with arbitrary node distributions.
\newblock {\it Transportation Science\/} {\bf 58}(1) 45--66.

\bibitem[{Vaswani et~al.(2017)Vaswani, Shazeer, Parmar, Uszkoreit, Jones,
  Gomez, Kaiser, and Polosukhin}]{vaswani2017attention}
Vaswani, A., N. Shazeer, N. Parmar, J. Uszkoreit, L. Jones, A.~N. Gomez, {\L}.
  Kaiser, I. Polosukhin. 2017.
\newblock Attention is all you need.
\newblock {\it Advances in Neural Information Processing Systems\/} {\bf 30}.

\bibitem[{Vidal(2022)}]{vidal2022hybrid}
Vidal, T. 2022.
\newblock Hybrid genetic search for the {CVRP}: Open-source implementation and
  {SWAP}* neighborhood.
\newblock {\it Computers \& Operations Research\/} {\bf 140} 105643.

\bibitem[{Vidal et~al.(2012)Vidal, Crainic, Gendreau, Lahrichi, and
  Rei}]{vidal2012hybrid}
Vidal, T., T.~G. Crainic, M. Gendreau, N. Lahrichi, W. Rei. 2012.
\newblock A hybrid genetic algorithm for multidepot and periodic vehicle
  routing problems.
\newblock {\it Operations Research\/} {\bf 60}(3) 611--624.

\bibitem[{Vidal et~al.(2014)Vidal, Crainic, Gendreau, and
  Prins}]{vidal2014implicit}
Vidal, T., T.~G. Crainic, M. Gendreau, C. Prins. 2014.
\newblock Implicit depot assignments and rotations in vehicle routing
  heuristics.
\newblock {\it European Journal of Operational Research\/} {\bf 237}(1) 15--28.

\bibitem[{Vinyals et~al.(2015)Vinyals, Fortunato, and
  Jaitly}]{vinyals2015pointer}
Vinyals, O., M. Fortunato, N. Jaitly. 2015.
\newblock Pointer networks.
\newblock {\it Proceedings of the 28th International Conference on Neural
  Information Processing Systems-Volume 2\/}. 2692--2700.

\bibitem[{Wu et~al.(2021)Wu, Song, Cao, Zhang, and Lim}]{wu2021learning}
Wu, Y., W. Song, Z. Cao, J. Zhang, A. Lim. 2021.
\newblock Learning improvement heuristics for solving routing problems.
\newblock {\it IEEE Transactions on Neural Networks and Learning Systems\/}
  {\bf 33}(9) 5057--5069.

\bibitem[{Xu et~al.(2020)Xu, Zhang, Li, Du, Kawarabayashi, and
  Jegelka}]{xu2020neural}
Xu, K., M. Zhang, J. Li, S.~S. Du, K.-i. Kawarabayashi, S. Jegelka. 2020.
\newblock How neural networks extrapolate: From feedforward to graph neural
  networks.
\newblock {\it arXiv preprint arXiv:2009.11848\/} .

\bibitem[{Zhang et~al.(2023)Zhang, Lin, and Li}]{zhang2023graph}
Zhang, K., X. Lin, M. Li. 2023.
\newblock Graph attention reinforcement learning with flexible matching
  policies for multi-depot vehicle routing problems.
\newblock {\it Physica A: Statistical Mechanics and its Applications\/}
  128451.

\bibitem[{Zong et~al.(2022)Zong, Wang, Wang, Zheng, and Li}]{zong2022rbg}
Zong, Z., H. Wang, J. Wang, M. Zheng, Y. Li. 2022.
\newblock Rbg: Hierarchically solving large-scale routing problems in logistic
  systems via reinforcement learning.
\newblock {\it Proceedings of the 28th ACM SIGKDD Conference on Knowledge
  Discovery and Data Mining\/}. 4648--4658.

\bibitem[{Zou et~al.(2022)Zou, Wu, Yin, Dhamotharan, Chen, and
  Tiwari}]{zou2022improved}
Zou, Y., H. Wu, Y. Yin, L. Dhamotharan, D. Chen, A.~K. Tiwari. 2022.
\newblock An improved transformer model with multi-head attention and attention
  to attention for low-carbon multi-depot vehicle routing problem.
\newblock {\it Annals of Operations Research\/}  1--20.

\end{thebibliography}
